\documentclass[]{amap}

\usepackage[toc,page,header]{appendix}
\usepackage{minitoc}
\usepackage{solarized-light}
\usepackage{booktabs}
\usepackage{multirow}
\usepackage{graphicx}
\usepackage{array}
\usepackage{siunitx,array}
\usepackage[table]{xcolor}
\usepackage{makecell,array,booktabs}
\usepackage{makecell}
\usepackage{framed}
\usepackage{amssymb}
\usepackage{bbding}
\usepackage{adjustbox}
\usepackage{amsmath}
\usepackage{placeins}
\usepackage[colorinlistoftodos]{todonotes}
\usepackage{longtable}
\usepackage{hhline}
\usepackage{fancyvrb}
\usepackage{float}
\usepackage{fvextra}
\usepackage{CJKutf8}
\usepackage{multicol}
\usepackage{tablefootnote}
\usepackage{threeparttable}
\usepackage{tabularx}
\usepackage{mdframed}
\usepackage{subcaption}
\usepackage[usestackEOL]{stackengine}
\usepackage{enumitem}
\usepackage{microtype}
\usepackage{url}
\usepackage{amsthm}

\newcommand{\commentout}[1]{}
\renewcommand{\paragraph}[1]{\noindent\textbf{#1}\hspace*{1em}}
\setlist[itemize]{leftmargin=15pt}

\RequirePackage{xspace}
\makeatletter
\DeclareRobustCommand\onedot{\futurelet\@let@token\@onedot}
\def\@onedot{\ifx\@let@token.\else.\null\fi\xspace}

\makeatother

\newcommand{\MODEL}{ABot-N1\xspace}

\graphicspath{{figs/}}

\title{\MODEL: Toward a General Visual Language Navigation Foundation Model}

\vspace{-10pt}

\abstract{
Visual Language Navigation foundation models aim to unify deep \textit{reasoning} for grounded spatial decisions with broad \textit{versatility} for diverse embodied tasks. 
Current approaches typically achieve this integration via monolithic policies that map observations directly to actions, yet they often suffer from coordinate drift and poor handling of long-tail semantics. 
Furthermore, these black-box mappings lack interpretability, hindering the simultaneous achievement of generality, robustness, and transparency. 
We present \textbf{ABot-N1}, a step toward a general Visual Language Navigation foundation model, that addresses these challenges by decoupling cognition from control via a \textbf{slow-fast} architecture guided by dual visual-language signals.
More specifically, a slow vision-language reasoner performs explicit Chain-of-Thought \textbf{reasoning} while producing a \textbf{pixel goal}. This compact set of image-space anchor points serves as a universal interface for diverse tasks, including \textit{point-goal}, \textit{object-goal}, \textit{poi-goal}, \textit{instruction-following}, and \textit{person-following}.
Subsequently, a fast action expert leverages both the textual cues and the pixel guidance to generate continuous waypoints at the native control frequency. 
By bridging high-level intents and low-level control through pixel-grounded anchors paired with explicit linguistic traces, our approach ensures robust, generalizable, and interpretable navigation across simulation and real-world benchmarks.
ABot-N1 establishes new state-of-the-art records, delivering massive gains specifically in urban-scale navigation: \textbf{boosting POI arrival by 35.0\% (to 77.3\%) and achieving 95.4\%/92.9\% SR in complex indoor and outdoor scenes}. It also maintains superior robustness across object-reaching, person-following,  and instruction-following tasks. 
New Point-Goal/POI-Goal benchmarks are released as open source to advance the field of urban-scale navigation.

\textbf{Project Page:} \url{https://amap-cvlab.github.io/ABot-Navigation/ABot-N1/}

}

\begin{document}
\maketitle
\vspace{-4pt}

{%
\vspace{-14pt}

\centering
\includegraphics[width=\linewidth,height=0.36\textheight,keepaspectratio]{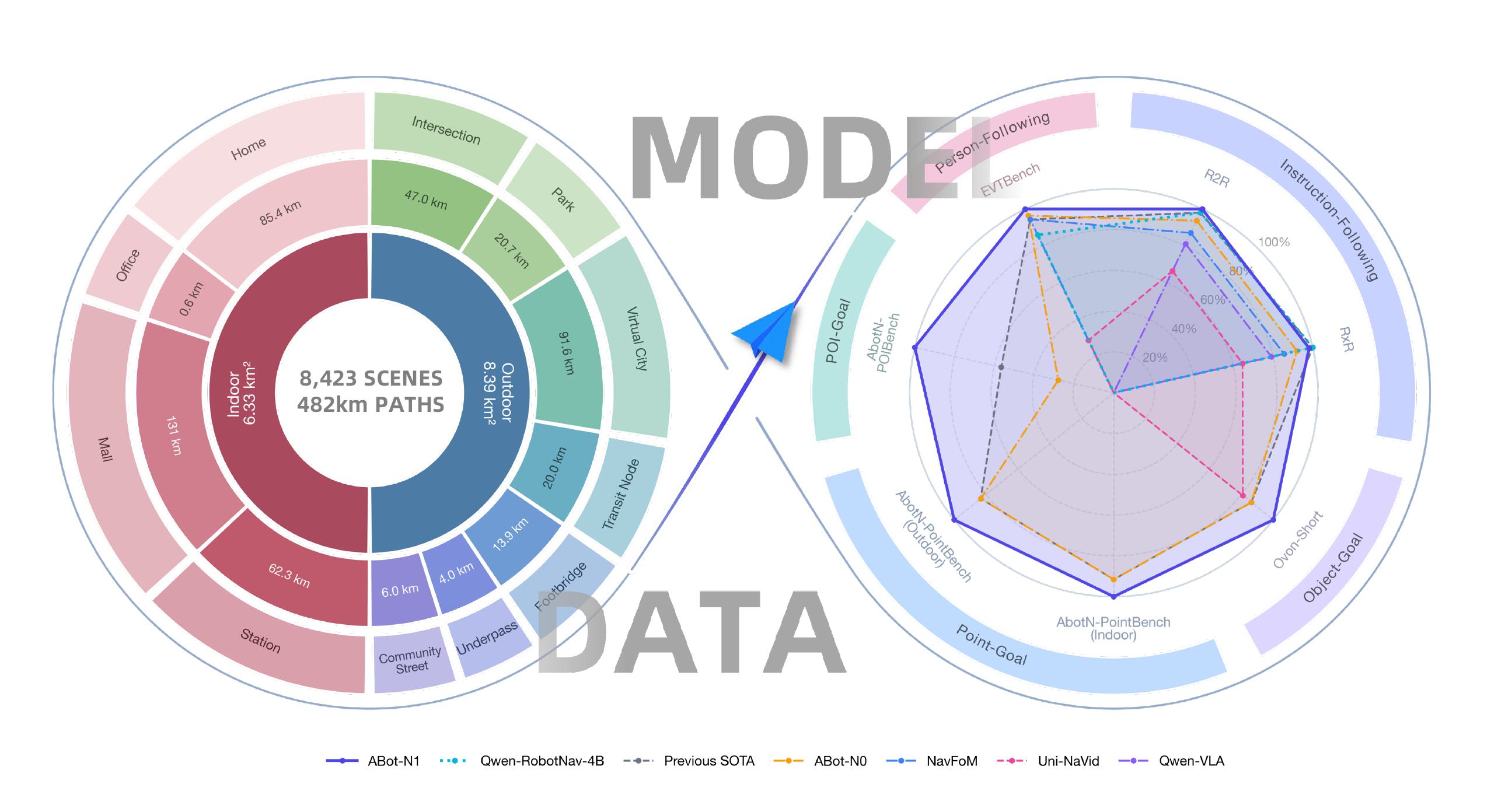}

\par}
\newpage
\tableofcontents
\newpage

\section{Introduction}
\label{sec:intro}

Embodied navigation serves as a fundamental capability for general-purpose physical agents, bridging high-level cognition and low-level motor control~\cite{anderson2018vln,gu2022vlnsurvey,hossain2026review,yang2025nav,gong2026poinavbenchmarkingenhancingfinalmeters,yang2026asyncshieldplugandplayedgeadapter,chen2026explorelikehumansautonomous,chu2026abotn0technicalreportvla,chen2026astranavworldworldmodelforesight,xiang2025navr2dualrelationreasoninggeneralizable,Liu_2026_CVPR,Chen_2026_CVPR,xue2026omninavunifiedframeworkprospective,yang2025cenavflowguidedreinforcementrefinement}. 
A truly versatile navigator must operate in open-world environments—where lighting, geometry, dynamics, and social norms vary continuously—to reach precise metric targets, follow free-form natural-language instructions, search for open-vocabulary objects, locate points of interest (POIs), and track moving people.
However, historically, these abilities have been pursued in isolation. 
Tasks such as \textit{point-goal}, \textit{object-goal}, \textit{POI-goal}, \textit{instruction-following}, and \textit{person-following} are typically dominated by task-specific architectures characterized by bespoke goal interfaces, narrow training data, and limited cross-task transfer~\cite{anderson2018vln,batra2020objectnav,krantz2020vlnce,savva2019habitat}.
This fragmentation hinders the extraction of physical and semantic priors from heterogeneous data, resulting in a ``\textit{zoo}'' of specialized models that are difficult to integrate into a single robot.

Recent efforts address this fragmentation by developing general-purpose navigation foundation models~\cite{zhang2025navfom,abotn0,sun2026openfrontier,chen2025socialnav} that unify diverse tasks within a single architecture and corpus. 
While this paradigm has demonstrated improved cross-task transfer, three critical \textbf{challenges} persist as these generalist models transition from benchmark leaderboards to real-world robotic deployment:
\begin{itemize}

\item \textbf{To-Point Navigation.} Target coordinates are specified as ego-centric offsets relative to the robot’s pose. However, inaccuracies in Standard-Definition (SD) map routing or localization can shift these targets into physically infeasible regions—such as vehicle lanes or flower beds—instead of valid walkable paths. This misalignment necessitates explicit traversability reasoning to ensure safe arrival.


\item \textbf{To-Target Navigation.} End-to-end fine-tuning on object-search trajectories risks eroding pretrained semantic priors critical for open-vocabulary recognition. Moreover, conflating ``search'' and ``approach'' phases couples stochastic exploration with deterministic execution, leading to brittle training convergence and ambiguous failure attribution.

\item \textbf{Interpretability \& Safety.} Monolithic policies that map observations directly to actions operate as black-box systems, lacking explicit, human-readable decision traces or intermediate reasoning steps. This opacity obscures the causal link between sensory inputs and motor outputs, severely hindering root-cause analysis of failures and complicating safety audits, as developers cannot easily discern whether errors stem from perception, reasoning, or control modules.
\end{itemize}



These pain points share a fundamental architectural root cause: the reliance on monolithic policies that bypass intermediate reasoning. 
This design suffers from two critical flaws. 
Firstly, it creates a spectral mismatch between the slow dynamics of semantic reasoning and the fast dynamics of motor control, forcing distinct computational processes into a single update cycle. 
Secondly, optimizing for heterogeneous tasks within a shared parameter space leads to optimization interference, where conflicting gradients from diverse objectives (e.g., precision navigation vs. semantic exploration) cause negative transfer~\cite{yu2020gradient,liu2021conflict}. 
We argue that a modular, factorized approach offers a superior solution. 
By decoupling cognition from control, we enable a low-frequency reasoner to perform explicit planning and a high-frequency controller to handle reactive execution. These components interact through a transparent, structured bottleneck, replacing the black-box latent representations of traditional end-to-end models.

\begin{figure}[t]
\centering
\includegraphics[width=\linewidth]{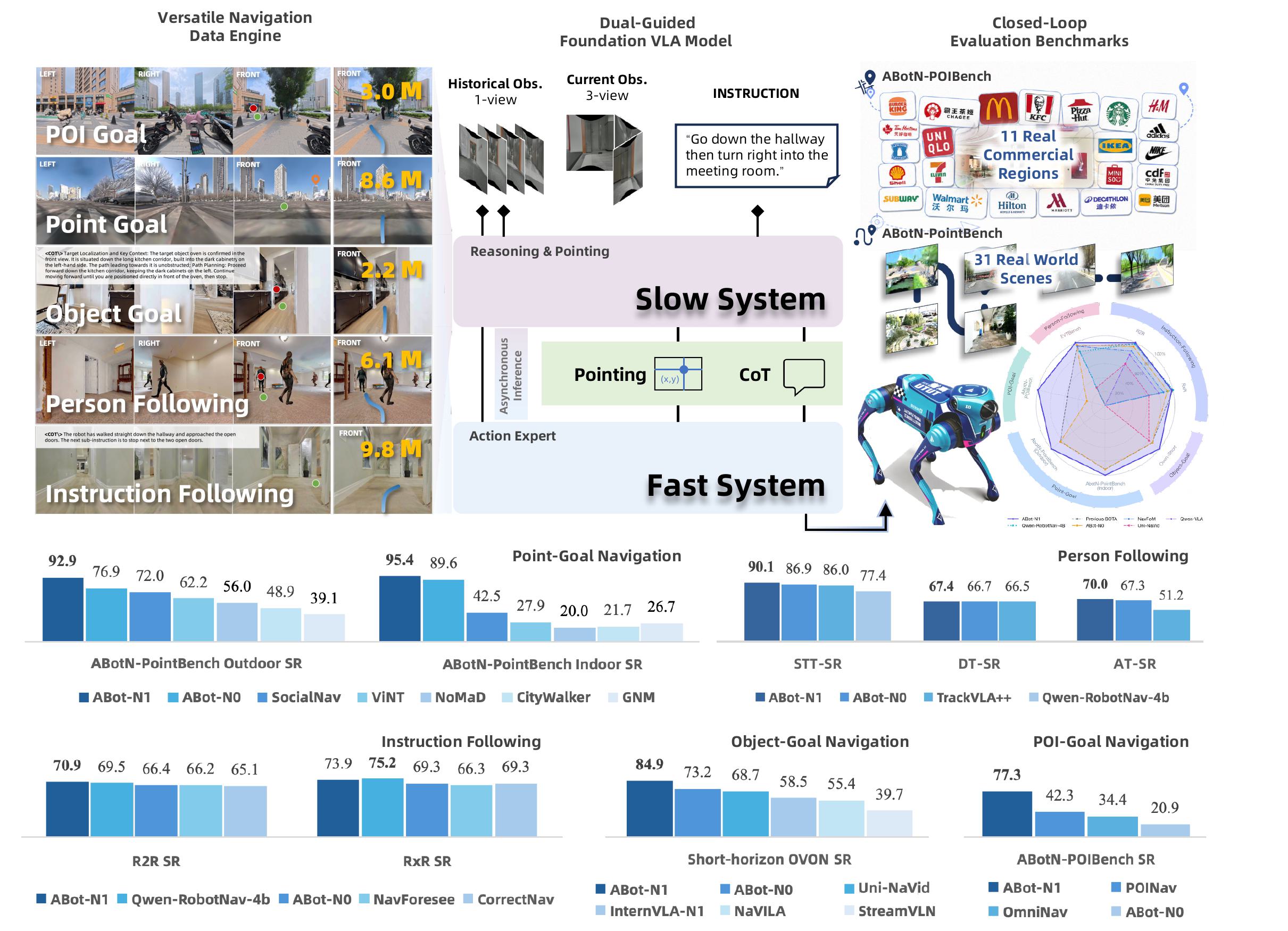}

\caption{\textbf{Overview of ABot-N1.} The model trained on 30M samples across five tasks adopts a slow--fast control architecture: a slow system performs CoT reasoning and emits pixel goals, while a fast action expert consumes this dual language-and-vision guidance to execute safe waypoints.
Closed-loop evaluation is conducted on our newly proposed ABotN-PointBench and ABotN-POIBench, together with three established benchmarks (VLN-CE R2R/RxR, Short-Horizon OVON, and EVT-Bench). ABot-N1 achieves leading performance across all 5 benchmarks.}
\label{fig:overview}
\end{figure}

Motivated by these insights, we introduce \textbf{ABot-N1} (see Figure~\ref{fig:overview}), a Vision-Language-Navigation (VLN) foundation model featuring a hierarchical \textbf{slow–fast} architecture. The slow subsystem employs a 4B-parameter multimodal VLM to conduct high-level deliberative reasoning. By processing instructions, visual memory, and current observations, it produces a CoT rationale that interprets intent, grounds language to visual entities, and adheres to social norms ~\cite{wei2022cot,deepseekai2025r1,chen2025socialnav}. Crucially, this reasoning process yields a pixel goal—a compact set of 2D anchors in the egocentric view that defines the next sub-goal. The fast subsystem, a specialized 2B-parameter VLM, acts as a low-latency action expert. It synthesizes the slow system’s CoT and pixel goal with real-time visual inputs to emit continuous waypoints, ensuring responsive control at the robot’s native frequency.

The core design principle of our architecture lies in its \textbf{dual-modality guidance}: the slow system emits a CoT for semantic reasoning and a Pixel Goal for spatial grounding. 
The linguistic traces provide robust, transferable high-level logic, while the pixel goal translates this logic into actionable, egocentric visual anchors. 
Regardless of the initial task formulation---whether a metric coordinate, a free-form instruction, an object category, a POI name, or a target person---it is decomposed into this shared language-plus-vision representation.
By reducing all tasks to \textit{``tracking CoT-explained pixels''}, the fast controller operates on a consistent interface regardless of input complexity. 
This approach generalizes earlier visual-prompting techniques~\cite{feng2026vpn,nasiriany2024pivot,sun2026openfrontier} by evolving from passive, user-defined prompts to active, reasoning-aware model generations. 
As a result, the framework seamlessly handles geometric, semantic, and dynamic targets. 
Three distinct \textbf{advantages} emerge from this design:
\begin{itemize}

\item \textbf{To-Point Navigation.} By performing image-space re-grounding at each step, we mitigate the sensitivity to coordinate inaccuracies inherent in ego-centric offset targets. When localization or map errors shift the goal into non-traversable areas, the CoT-guided affordance pixels automatically correct the trajectory toward the nearest safe pathway, preventing the agent from getting stranded in prohibited regions—a common failure mode for metric-only baselines.

\item \textbf{To-Target Navigation.} By offloading open-vocabulary recognition to the deliberative slow reasoner, we decouple semantic understanding from reactive control. The slow system utilizes its robust pre-trained priors to resolve ambiguities in long-tail distributions and compositional queries, which are typically prone to failure in end-to-end policies. This separation ensures that complex semantic interpretations do not compromise the stability of the fast controller, leading to more reliable identification of novel and intricate targets.

\item \textbf{Interpretability \& Safety.} By integrating human-readable CoT reasoning with visually grounded pixel anchors, our architecture achieves fine-grained interpretability at every decision step. This transparency is crucial for safety-critical applications, as it allows for precise root-cause analysis of failures. Developers can distinguish whether an error stems from flawed semantic reasoning (visible in the CoT) or inaccurate spatial targeting (visible in the pixel goal), thereby accelerating debugging and ensuring that the robot’s behavior aligns with human expectations and safety constraints.
\end{itemize}

Realizing this design at scale requires overcoming the data and optimization bottlenecks prevalent in existing generalist models. We address this by extending our prior data infrastructure~\cite{abotn0} with high-fidelity, pixel-grounded reasoning traces for all target tasks. Crucially, we depart from standard token-level supervision by introducing a GRPO-based reinforcement learning stage~\cite{deepseekai2025r1} for the slow system. This mechanism aligns the model’s output with actual task completion rather than just linguistic plausibility, thereby enhancing the robustness and physical feasibility of the generated pixel goals.

Despite the maturity of benchmarks for \textit{instruction-following} and \textit{object-goal} tasks~\cite{anderson2018vln,ku2020rxr,krantz2020vlnce,yokoyama2024hmovon}, open-source evaluation frameworks for cityscale navigation—specifically \textit{point-goal} and \textit{POI-goal}—remain notably scarce. 
We bridge this gap by releasing two novel benchmarks: \textbf{ABotN-PointBench} covering complex indoor and outdoor environments, and \textbf{ABotN-POIBench} situated in realistic commercial districts. 
Through a hierarchical difficulty structure, these benchmarks enable comprehensive assessment of state-of-the-art methods and demonstrate the superior capabilities of ABot-N1. 
We make both benchmarks publicly available to foster further advancements in the field.

\textbf{In summary, this report makes the following contributions:}
\begin{itemize}
  \item \textbf{Architecture.} We introduce ABot-N1, a dual-system VLA navigator featuring a slow–fast architecture and a unified pixel-goal interface. It provides joint semantic-spatial guidance via CoT and image anchors for five key tasks: \textit{point-goal}, \textit{object-goal}, \textit{POI-goal}, \textit{instruction-following}, and \textit{person-following}.
  
  \item \textbf{Training Recipe.} Building on the ABot-N0 data engine~\cite{abotn0}, we integrate pixel-grounded supervision and CoT trajectories for all five tasks. Furthermore, we employ GRPO-style post-training~\cite{deepseekai2025r1} to align the slow system's reasoning directly with downstream navigation rewards.
  
  \item \textbf{Cityscale Benchmarks.} We release ABotN-PointBench and ABotN-POIBench, which comprehensively evaluate \textit{point-goal} navigation (short/long-range, indoor/outdoor) and \textit{POI-goal} navigation in real-world commercial scenarios.
  
  \item \textbf{Cityscale Navigation.} Using a unified checkpoint, ABot-N1 achieves state-of-the-art performance across five tasks, exceeding both specialized and multitask models. It further propels urban navigation forward by enabling robust long-horizon autonomy in complex cityscapes with only low-fidelity SD maps, thereby overcoming the traditional reliance on high-precision infrastructure.
  
\end{itemize}

More intuitively, this slow-fast separation allows us to treat semantic recognition and motor action as distinct data streams. By decoupling the slow reasoning system from the fast control system, we effectively disentangle these domains—a distinction crucial for open-world scaling as it enables the independent expansion of training corpora for perception and control without interference. Having already validated the effectiveness of this paradigm in \textbf{ABot-N1}, we plan to further verify the efficacy of this data-centric scaling strategy in our next iteration, \textbf{ABot-N1.1}.
\section{Related Works}
\label{sec:related}

We organize prior work along three axes that are most relevant to ABot-N1:
(i) general-purpose navigation foundation models that move from
single-task specialists to unified policies, (ii) the emerging
``brain--body'' decoupling represented by dual-system VLN architectures
that separate deliberation from reactive control, and (iii) the rise of
reasoning---both linguistic and visual---inside navigation, which we view
as a stepping stone toward general embodied reasoning models. Comprehensive
treatments of the broader VLN landscape can be found in recent
surveys~\cite{gu2022vlnsurvey,hossain2026review,zhang2026llmvln}.

\subsection{Generalist Navigation Foundation Models}
\label{sec:rw-generalist}

Classical VLN research is dominated by task-specific architectures that
specialize either the goal interface or the policy backbone for a single
benchmark, spanning \textit{instruction-following} on R2R/RxR-style
graphs~\cite{anderson2018vln,ku2020rxr,chen2021topological,an2024etpnav,an2022rxrhabitat,chen2022weakly},
open-vocabulary object goal
search~\cite{yokoyama2024hmovon,batra2020objectnav}, and geometric
\textit{point-goal} or \textit{person-following}
controllers~\cite{savva2019habitat,wang2025trackvla}. These specialists
rarely transfer across tasks and adopt disjoint goal interfaces that
resist joint training.

A recent line of work instead consolidates multiple navigation tasks
under a single video-conditioned VLM-based
policy~\cite{zhang2024navid,zhang2024uninavid,chiang2024mobilityvla},
while larger-scale foundation models such as
NavFoM~\cite{zhang2025navfom} and our prior
ABot-N0~\cite{abotn0} push toward cross-embodiment, multi-task coverage,
and training-free routes such as OpenFrontier~\cite{sun2026openfrontier}
ground language to visual frontiers for open-world generalization.
ABot-N1 inherits this commitment to a single, multi-task architecture
but departs from prior generalists in two ways: it explicitly factorizes
deliberation from control rather than emitting actions from a monolithic
network, and it routes all tasks through a shared \emph{pixel-goal}
interface that the fast controller can chase regardless of the original
goal modality.

\subsection{Brain-Body Decoupling: Dual-System VLN Architectures}
\label{sec:rw-dualsystem}

A central trend across robotics, autonomous driving, and humanoid control
is the explicit separation of a high-capacity ``big brain'' that
deliberates over multimodal context from a lightweight ``small brain''
that executes high-frequency motor commands. This brain--body division
of labor is motivated by the very different time constants and
uncertainty profiles of cognition and control, and it is most often
framed in the language of Kahneman's dual-process theory: a slow,
deliberative System~2 paired with a fast, reactive System~1.

The pattern has crystallized across domains: VLM-based deliberators paired
with fast low-level controllers in autonomous
driving~\cite{wen2024drivevlm}, hierarchical or asynchronously coupled
VLAs in manipulation and humanoid
control~\cite{shi2025hi,figureai2025helix,physicalintelligence2024pi05},
and dedicated humanoid foundation models that make the System~1/System~2
split explicit, most notably NVIDIA's GR00T~N1~\cite{nvidia2025groot}.
Follow-up work refines the interface between the two systems with
value-guided thinking, shared-representation execution, or unified
embodied brains~\cite{song2025hume,chen2025fastinslow,baai2025robobrain2},
and the same brain--body split has been imported into navigation for
socially-aware behavior~\cite{chen2025socialnav}.

ABot-N1 inherits this brain--body philosophy but specializes it for
embodied navigation in two important respects. First, the interface
between the two systems is \emph{not} an opaque latent vector or a free-form
natural-language sub-goal, but a structured \emph{pixel goal} accompanied
by a Chain-of-Thought explanation; this bottleneck is human-readable,
imitation-friendly, and amenable to reinforcement learning at the
reasoning level. Second, the slow system is post-trained with
GRPO~\cite{deepseekai2025r1} so that its reasoning is supervised by
downstream navigation outcomes rather than only by token-level targets.

\subsection{Toward General Embodied Reasoning for Navigation}
\label{sec:rw-reasoning}

Orthogonal to the architectural question of how to split brain and body
is the representational question of \emph{what} the deliberative side
should reason about and how that reasoning should be expressed. Recent
work has converged on two complementary answers: explicit linguistic
CoT, and structured visual representations such as visual
prompts and grounded spatial reasoning. We see ABot-N1's CoT-plus-pixel-goal
output as an instance of this broader trend toward general embodied
reasoning.

On the linguistic side, CoT prompting and its
reinforcement-learning descendants have shown that step-by-step textual
reasoning markedly improves compositional decision
making~\cite{wei2022cot,deepseekai2025r1}, and this idea has been
instantiated inside navigation through navigational chain-of-thought
training that disentangles reasoning from action
prediction~\cite{zhou2024navgpt2,lin2025navcot,zuo2026fantasyvln}, with
parallel embodied chain-of-thought recipes in
manipulation~\cite{zawalski2024ecot,sun2025emmax}.

On the visual side, a parallel thread argues that purely textual
reasoning under-uses the structure of the scene and grounds reasoning in
image-space artifacts---numbered action proposals, user-drawn goals,
frontier marks, or predicted subgoal
images~\cite{nasiriany2024pivot,feng2026vpn,sun2026openfrontier,zhao2025cotvla}.
A complementary line invests in the spatial competence of the underlying
VLM through quantitative and region-grounded 3D reasoning and unified
embodied perception~\cite{chen2024spatialvlm,cheng2024spatialrgpt,baai2026robobrain25}.

ABot-N1 unifies these two threads inside a single navigator. The slow
system produces an explicit Chain-of-Thought in the spirit of NavCoT,
NavGPT-2, ECoT, and FantasyVLN, but, in line with PIVOT, VPN, CoT-VLA,
and SpatialVLM, the reasoning grounds out in image-space \emph{pixel
goals} rather than text-only sub-goals or category embeddings. This
generalizes visual prompting beyond user-drawn marks and beyond
manipulation, and it makes the slow system's behavior directly
amenable to outcome-driven post-training. We view this combination as a
concrete step toward general embodied reasoning models, in which a
shared deliberative core can be specialized into many physical tasks by
what it points at on the image, while the fast controller solves the
universal problem of ``follow the predicted pixels.''

\section{Preliminaries}
\label{sec:preliminaries}

Versatile embodied navigation requires a single agent to fulfill
heterogeneous user goals---ranging from precise metric coordinates to
abstract semantic descriptions---using only on-board visual observations
and producing executable motion commands at the control rate. Before
describing ABot-N1, we cast this requirement as a unified
conditional sequential decision problem, specialize the goal
specification to the five tasks considered in this work, and introduce
the VLM-conditioned action-policy abstraction that subsequent sections
build on.

\subsection{Embodied Navigation As Goal-Conditioned Visual Control}
\label{sec:prelim-formulation}

We model an embodied navigator as a discrete-time agent acting in an
unknown environment $\mathcal{E}$. At each step $t$, the agent receives
an ego-centric RGB observation
$I_t \in \mathbb{R}^{N\times H\times W\times 3}$ from $N$ cameras and an
optional proprioceptive/odometry signal $q_t$, and is conditioned on a
task specification $g$ that encodes the user's intent. Following recent
navigation foundation
models~\cite{zhang2025navfom,zhang2024uninavid,physicalintelligence2024pi05},
we formulate the navigator as a goal-conditioned visual policy
\begin{equation}
\pi_\theta:\;\bigl(g,\,I_{1:t},\,q_{1:t}\bigr)\;\longmapsto\; a_{t:t+H}
\label{eq:nav-policy}
\end{equation}
which predicts a chunk of $H$ future low-level commands
$a_{t:t+H}=(a_t,a_{t+1},\ldots,a_{t+H-1})$ from the visual history
$I_{1:t}$ and the goal condition $g$. Each command
\begin{equation}
a_i=\bigl(x_i,\,y_i,\,\sin\theta_i,\,\cos\theta_i,\,c_i\bigr)\in\mathbb{R}^4\times\{0,1\}
\label{eq:waypoint}
\end{equation}
is a continuous $\mathrm{SE}(2)$ waypoint in the agent's local frame
augmented with a binary completion flag $c_i$ that triggers the
``arrive'' signal once the goal is reached. The agent rolls out
$\pi_\theta$ in a closed loop, re-planning at every step from updated
observations, until $c_i=1$ or a step budget is exhausted; an episode
succeeds when the final pose lies within a task-specific tolerance of
the true target. Crucially, this formulation is map-free: $\pi_\theta$
is given neither a global metric map nor depth or semantic
segmentation, and must extract all necessary geometry and semantics
from the on-the-fly RGB stream alone, matching the constraints of
real deployment on commodity
sensors~\cite{zhang2024navid,zhang2024uninavid}.

\subsection{Unifying FIVE Navigation Tasks in ONE Framework}
\label{sec:prelim-tasks}

The goal condition $g$ is realized as a natural-language instruction
$g=\ell$, where $\ell$ is a free-form sentence that encodes the
navigation intent. This single interface absorbs
all five tasks studied in this report without per-task heads or
bespoke encoders: in \textit{point-goal}, $\ell$ encodes a relative
metric offset; in \textit{instruction-following}
(R2R/RxR/VLN-CE-style episodes), $\ell$ is a step-by-step natural-language
route description; in \textit{object-goal}, $\ell$ is an open-vocabulary
category phrase such as ``find the nearest chair''; in \textit{POI-goal},
$\ell$ names a point of interest characterized by a landmark or
business identity; and in \textit{person-following}, $\ell$ describes
the target person via attribute or identity clauses.
Casting heterogeneous goal modalities into a unified language instruction
$\ell$ lets a single policy $\pi_\theta$ share parameters and
representations across tasks~\cite{zhang2025navfom,abotn0,zhang2024uninavid},
which is a prerequisite for transferring semantic priors learned on one
task to another.

\subsection{VLM-Conditioned Policy and Continuous-Action Decoding}
\label{sec:prelim-vlm}
Realizing $\pi_\theta$ at the scale required for open-world
generalization motivates initializing the policy from a pre-trained
vision-language
model~\cite{bai2023qwenvl,chen2024spatialvlm,physicalintelligence2024pi05}.
A natural design factorizes the navigator into a \emph{VLM reasoner}
$f_\phi$ and a lightweight \emph{action policy} $\psi_\xi$:
\begin{align}
s_t &= f_\phi\!\bigl(I_{1:t},\,\ell\bigr),
\label{eq:vlm-reason}\\
a_{t:t+H} &= \psi_\xi\!\bigl(s_t,\,I_{1:t},\ell,\,q_t\bigr)
\label{eq:action-head}
\end{align}
where $s_t$ denotes an intermediate guidance signal produced by the
reasoner--its concrete form (latent features, decoded language, visual
anchors, or a combination thereof) is a design choice deferred to
specific instantiations.
This factorization brings two benefits.
First, the broad linguistic and visual priors acquired during
web-scale VLM pre-training transfer directly into navigation,
providing open-vocabulary recognition, spatial-relation understanding,
and instruction parsing without task-specific
modules~\cite{bai2023qwenvl,chen2024spatialvlm,cheng2024spatialrgpt}.
Second, decoupling the heavy reasoner from the reactive action policy
allows them to operate at \emph{asymmetric computational
budgets}---a principle that ABot-N1 materializes as a slow-fast
dual-system architecture detailed in \S\ref{sec:method}.

\section{Methods}
\label{sec:method}

\subsection{Model Architecture}
\label{sec:method-arch}

\begin{figure*}[t]
  \centering
  \includegraphics[width=\textwidth]{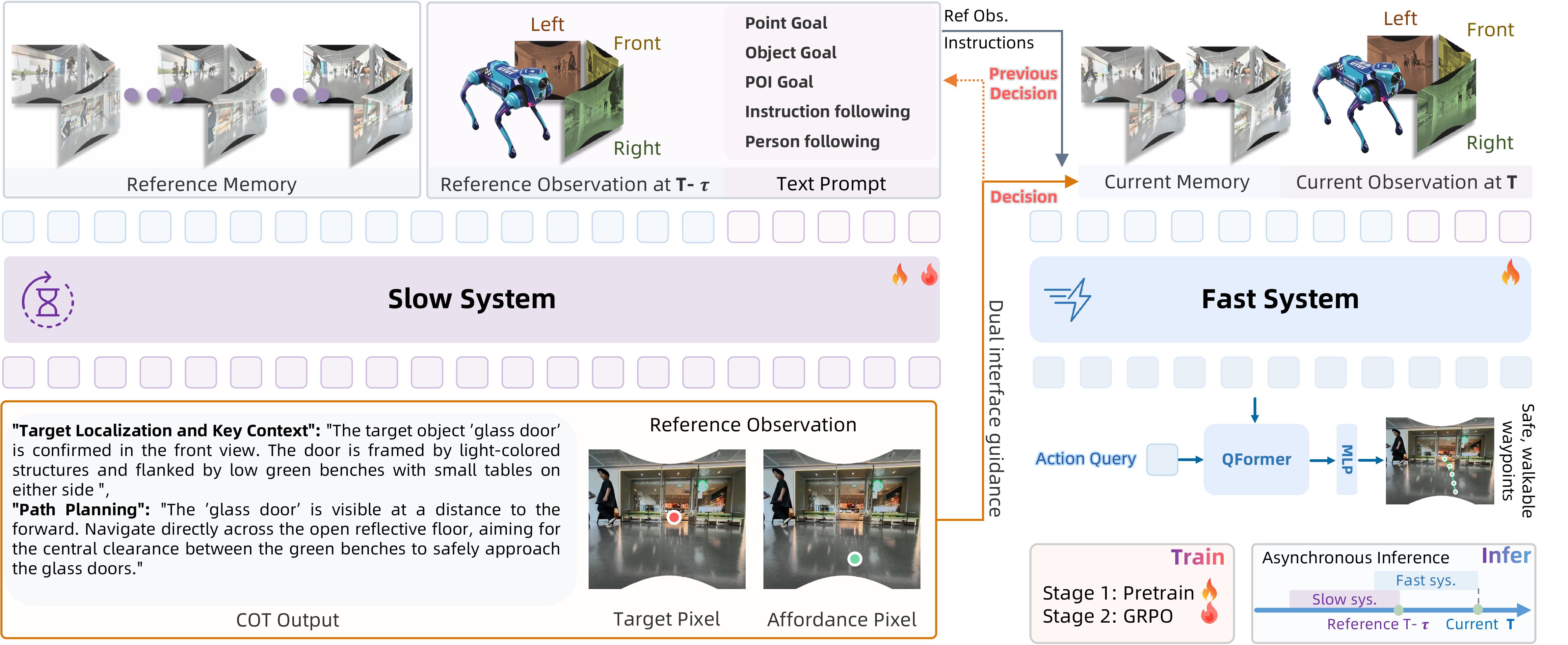}
  
  \caption{\textbf{The Slow-Fast Dual-System Architecture of ABot-N1.}
    Navigation is decoupled into asynchronous cognition and high-frequency control.
    Slow System (left): A vision-language reasoner processes historical frames
    and task prompts at low frequency, producing explicit CoT reasoning
    and visual anchors (Target Pixel and Affordance Pixel).
    Dual Vision-Language Interface (middle): The language and visual outputs
    form a unified bridge between the two systems.
    Fast System (right): A lightweight-VLM-based action expert integrates the dual guidance
    with real-time observations; a learnable action query attends to the output hidden states
    via a QFormer module, and an MLP decodes the queries to predict continuous waypoints.
    The system is trained with pretraining and GRPO, enabling complex reasoning without
    blocking the reactive control loop.}
  \label{fig:arch}
\end{figure*}

ABot-N1 decomposes the navigator of Eq.~\eqref{eq:nav-policy} into two
asynchronously coupled sub-networks: a \textit{slow system} that performs
deliberative reasoning and a \textit{fast system} that executes reactive
control. The two systems share the same visual front-end but differ
significantly in capacity, operating frequency, and responsibility.

\paragraph{Slow System - Deliberative Reasoner.}
The slow system is built on a 4B-parameter vision-language model
(Qwen-3.5-4B). Let $n\in\{1,2,\ldots\}$ index its successive
invocations.
At the $n$-th invocation it receives four categories of input:
(i)~a \textit{reference memory} $I^{\mathrm{ref,mem}}_{1:K}$---a short
buffer of historical front-view frames that summarizes the trajectory up
to the moment of this invocation;
(ii)~the \textit{reference observation} $I^{\mathrm{ref}}_{\mathrm{tri}}$---the
tri-view capture (left, front, right cameras) taken at this invocation;
(iii)~the task specification~$g=\ell$ defined in
\S\ref{sec:prelim-tasks}; and
(iv)~the \textit{previous decision} $(\mathcal{C}_{n-1},\,\mathbf{p}_{n-1})$
produced at invocation~$n{-}1$.
Feedback of the prior decision enables temporal consistency: for
tasks such as \textit{instruction-following} and \textit{person-following}, the model can
condition on its earlier decision to avoid contradictory or oscillating
behavior across successive inference cycles.
Given these inputs the slow system produces two complementary outputs:
\begin{equation}
f_\text{slow}\!\bigl(I^{\mathrm{ref,mem}}_{1:K},\, I^{\mathrm{ref}}_{\mathrm{tri}},\, g,\,
\mathcal{C}_{n-1},\,\mathbf{p}_{n-1}\bigr)
\;=\;\bigl(\mathcal{C}_n,\;\mathbf{p}_n\bigr)
\label{eq:slow}
\end{equation}
where $\mathcal{C}_n$ is an explicit natural-language Chain-of-Thought
trace that decomposes intent, grounds linguistic and target references, and $\mathbf{p}_n$ is a set of \textit{pixel
goals} projected onto the current tri-view images. The Chain-of-Thought
is emitted only for the semantically demanding tasks---\textit{instruction-following}
and \textit{object-goal}---whereas \textit{point-goal}, \textit{POI-goal}, and \textit{person-following}
rely on the pixel goals alone. The pixel goals
come in two semantic flavors, whose presence depends on the task and
the current state:
\begin{itemize}[nosep,leftmargin=1.2em]
\item \textbf{Affordance Pixel} An image-space point denoting where
  forward locomotion is currently afforded, i.e., the next safe,
  traversable waypoint the agent should pursue (typically
  $\sim$3\,m indoors or $\sim$5\,m outdoors ahead on the reachable
  ground).
\item \textbf{Target Pixel} An image-space point marking the final
  goal itself (e.g., an object, a POI entrance, or the tracked person),
  emitted only when the target becomes visible or the agent is within
  the final approach distance.
\end{itemize}
Different tasks instantiate these two slots in slightly different ways.
In \textit{point-goal}, the affordance pixel marks the next safe
waypoint on the occupancy-aware traversable area; no target pixel is
emitted (the metric coordinate goal is consumed by the slow system
internally and does not need an image-space anchor). In \textit{instruction-following}, the
affordance pixel corresponds to the near-future feasible waypoint along the
ground-truth path, and the target pixel is activated near the route
endpoint. In \textit{object-goal}, the affordance pixel guides exploration
while the target pixel marks the discovered object once recognized. In
\textit{POI-goal}, the affordance pixel follows the same convention as
\textit{point-goal}---marking the next safe traversable waypoint toward the
entrance---while the target pixel marks the physical entrance of the
queried POI once it is localized. In \textit{person-following}, both pixels
collapse to the bounding-box bottom center of the tracked person when
visible, or the affordance pixel falls back to the first projectable future
waypoint if the person temporarily exits the field of view.

\paragraph{Fast System - Action Expert.}
The fast system uses Qwen-3.5-2B as its backbone. It jointly encodes
four input streams: (i)~the \textit{current memory}
$I^{\mathrm{cur,mem}}_{1:K}$ and \textit{current observation}
$I^{\mathrm{cur}}_{\mathrm{tri}}$---both captured in real time,
(ii)~the slow system's CoT trace~$\mathcal{C}_n$ and pixel
goals~$\mathbf{p}_n$, (iii)~the slow system's reference observation
$I^{\mathrm{ref}}_{\mathrm{tri}}$---which anchors the pixel goals to their
original visual context---and (iv)~the task specification~$g$. The backbone
produces a fused hidden state
\begin{equation}
h_t = \mathrm{Enc}\!\bigl(
  I^{\mathrm{cur,mem}}_{1:K},\, I^{\mathrm{cur}}_{\mathrm{tri}},\,
  \mathcal{C}_n,\, \mathbf{p}_n,\, I^{\mathrm{ref}}_{\mathrm{tri}},\, g
\bigr)
\label{eq:fast-hidden}
\end{equation}
A set of learnable \textit{action queries} $q_{\mathrm{act}}$ then
interact with $h_t$ via a QFormer-style cross-attention module,
distilling navigation-relevant features into a compact query
representation. A lightweight MLP head decodes this representation
into the navigation output:
\begin{equation}
a_{t:t+H} = \mathrm{MLP}\!\bigl(\mathrm{QFormer}(q_{\mathrm{act}},\,h_t)\bigr)
\label{eq:fast}
\end{equation}
where $a_{t:t+H}$ is a sequence of $H{=}5$ continuous SE(2) actions
in the ego-centric frame (Eq.~\eqref{eq:waypoint}), each comprising
a 2-D position $(x_i, y_i)$, a heading angle encoded as
$(\sin\theta_i, \cos\theta_i)$, and a binary completion flag $c_i$,
with adaptive step sizes that are larger outdoors and smaller
indoors to balance efficiency and obstacle clearance.

\paragraph{Asynchronous Inference.}
The slow and fast systems operate at different cadences. The slow
system runs at a low frequency, limited by the
4B-parameter model's inference cost but sufficient for high-level
reasoning that changes on the timescale of navigation decisions. The
fast system runs at a higher rate, consuming the
most recent slow-system context to produce smooth, responsive
trajectories. Between successive slow-system updates, the fast system
continues to execute using the last available $(\mathcal{C}_t,
\mathbf{p}_t, I^{\mathrm{ref}}_{\mathrm{tri}})$, densely tracking the
cached pixel goal to bridge the temporal gap between sparse
deliberative updates. This asynchronous
coupling allows the slow system to reason deeply without stalling the
control loop, and lets the fast system compensate for the reasoning
latency through visual closed-loop tracking of the provided pixel
goals.

\begin{figure}[!t]
\centering
\includegraphics[width=\linewidth]{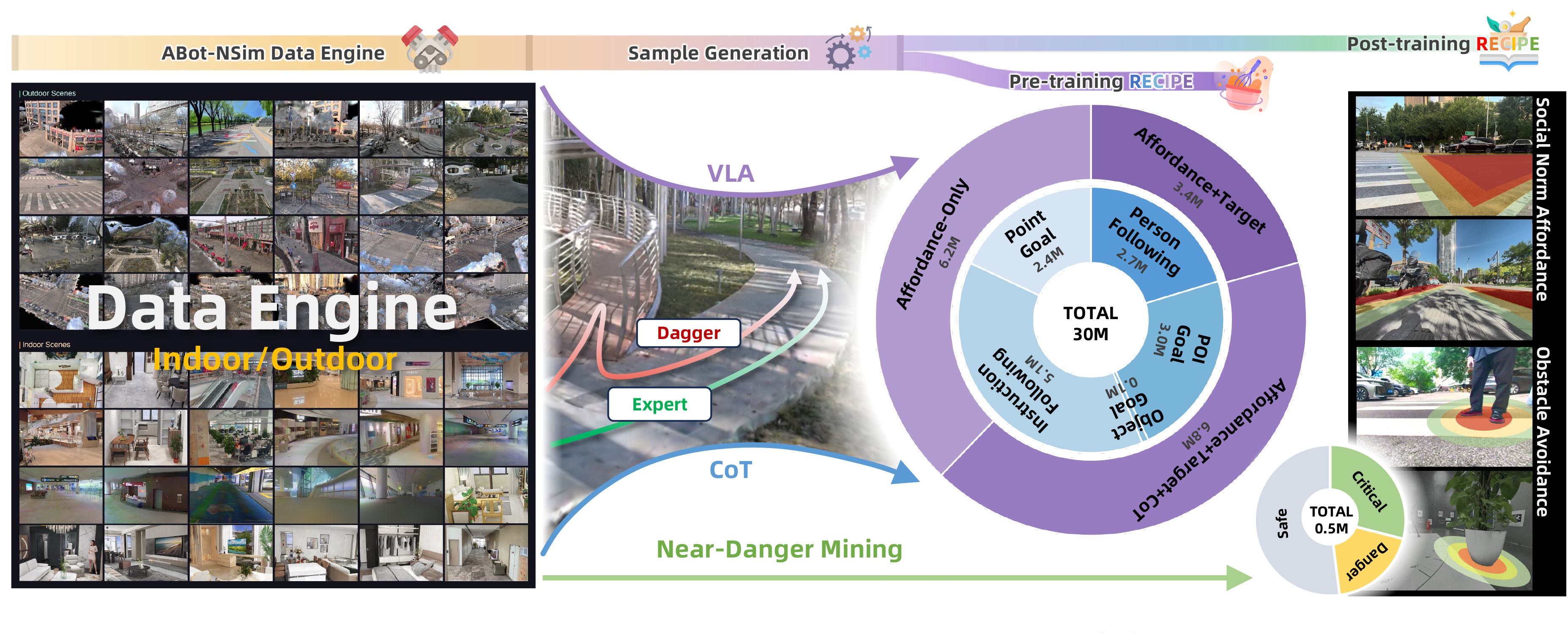}

\caption{\textbf{Data Pipeline and Composition.}
The data engine (left) provides diverse indoor and outdoor simulation scenes; trajectory generation (middle) produces expert and Dagger rollouts; the resulting samples (right) span both stages---the five pre-training navigation tasks broken down by slow-system (high-level) and fast-system (low-level) counts, together with the post-training composition stratified into Safe, Critical, Danger, and discarded data.}
\label{fig:data}
\end{figure}

\subsection{Pretraining}
\label{sec:method-pretrain}

Both the slow and fast systems are pre-trained via supervised
imitation on a heterogeneous corpus that we curate to cover all five
navigation tasks. Generalist behavior under Eq.~\eqref{eq:nav-policy} only
emerges when the training distribution covers the diversity of real
deployment; following recent navigation foundation
models~\cite{zhang2025navfom,abotn0,physicalintelligence2024pi05}, our
corpus spans photorealistic indoor simulation from Habitat-style scene
banks~\cite{savva2019habitat,batra2020objectnav,krantz2020vlnce},
outdoor driving and street-view trajectories, 
and real-robot teleoperation. We reformat each episode into the
$(g,\,I_{1:T},\,a_{1:T})$ tuple of Eq.~\eqref{eq:nav-policy} and
enrich it with chain-of-thought rationales and pixel-goal annotations.
The two systems are trained independently: the slow system learns to
produce CoT traces and pixel goals, while the fast system learns the
continuous-waypoint policy conditioned on slow-system outputs. We
first describe the general training principles, then summarize
task-specific characteristics.

\subsubsection{Training Protocol}
\label{sec:data-principles}

We initialize both systems from their respective pre-trained
Qwen-3.5 checkpoints. The slow system is supervised with a cross-entropy
language-modeling loss for CoT tokens and pixel coordinates, while the
fast system is supervised with a smooth-$L_1$ loss on position and
heading angle together with a binary arriving loss.  In generic VLN settings, a smooth-$L_1$ regression objective is
well known to induce \textit{mode collapse}: when the target
distribution is multi-modal (e.g., go left vs.\ go right with
comparable likelihood), the loss is minimized by the conditional
mean, which often corresponds to an averaged trajectory that is
invalid in either mode and causes severe performance degradation
on the downstream task. This is the standard motivation for GMM
heads, discretized action tokens, and flow- or diffusion-based
action heads. We can nevertheless rely on smooth-$L_1$ in
ABot-N1 because of the dual-system factorization: the slow
system has already committed to a discrete decision through its
CoT trace and pixel-goal output (target pixel + affordance
pixel), which together sharply concentrate the future-waypoint
distribution around a dominant mode. Conditioned on these
slow-system outputs, the fast policy faces an approximately
uni-modal regression problem, in which the conditional mean
closely approximates the desired trajectory and smooth-$L_1$
regression is both stable and sample-efficient. In other words,
the slow system substantially reduces the modal-selection burden
that would otherwise be handled by a heavier flow- or
diffusion-based action head.
 A shared design principle across
tasks is \textit{robustness augmentation}: we feed both systems
noise-injected inputs (perturbed target coordinates, noisy prior
pixel predictions, or simulated asynchronous lag) so that they learn
to tolerate imperfect upstream signals encountered during deployment.

\subsubsection{Pretraining Data Recipe}
\label{sec:data-tasks}

Figure~\ref{fig:data} summarizes the overall data pipeline and
composition across both training stages; here we focus on its
pre-training portion, while the post-training composition is detailed
in Section~\ref{sec:grpo-trainingdata}.

Our data engine maintains a bank of high-fidelity simulation
scenes spanning 800+ indoor and 22 outdoor environments reconstructed
as 3D Gaussian Splatting. From these scenes, trajectory generation
 produces two complementary types of rollouts: expert
demonstrations that provide strictly correct supervision, and Dagger
rollouts that introduce controlled errors and recoveries to build
corrective capabilities.
The resulting 30\,M pre-training samples  are distributed
across five navigation tasks--- \textit{point-goal},
\textit{object-goal}, \textit{POI-goal}, \textit{instruction-following}, and \textit{person-following}---organized into
high-level data (13.3\,M) for the slow system and low-level data
(16.4\,M) for the fast system. The low-level inputs come in three
configurations:
(1)~CoT + affordance pixel + target pixel for \textit{instruction-following}
and \textit{object-goal},
(2)~affordance pixel only for \textit{point-goal}, and
(3)~affordance pixel + target pixel for \textit{POI-goal} and \textit{person-following}.
Below we describe the data sources and construction pipeline for each task.

\paragraph{Point-Goal.}
Building upon the data engine of ABot-N0, we construct a unified scene bank of 800+ indoor and 22 outdoor scenes
reconstructed as high-fidelity 3D Gaussian Splatting (3DGS)
environments and manually annotate every scene with an occupancy map
that delineates traversable from prohibited regions (sidewalks,
crosswalks, building entrances vs.\ vehicle lanes, lawns, and static
obstacles). Episodes are sampled along curated trajectories: for each
starting pose we draw a target 5--50\,m away outdoors (3--20\,m
indoors), and the ground-truth affordance pixel is obtained by
selecting a safe traversable waypoint along the path (5\,m outdoors,
3\,m indoors), running a \textit{safety check}, and projecting
the resulting 3D point onto the current image plane. The safety check
verifies that the line of sight from the viewpoint to the candidate
point does not cross any non-traversable region; if it does, we
backtrack along the trajectory until a valid point that is both
unoccluded and traversable is found. From this shared
episode pool, the slow and fast systems derive their respective
training samples---often from the same trajectory sampling points but
with different input and supervision content, as summarized below.

\begin{figure*}[!t]
\centering
\includegraphics[width=\linewidth]{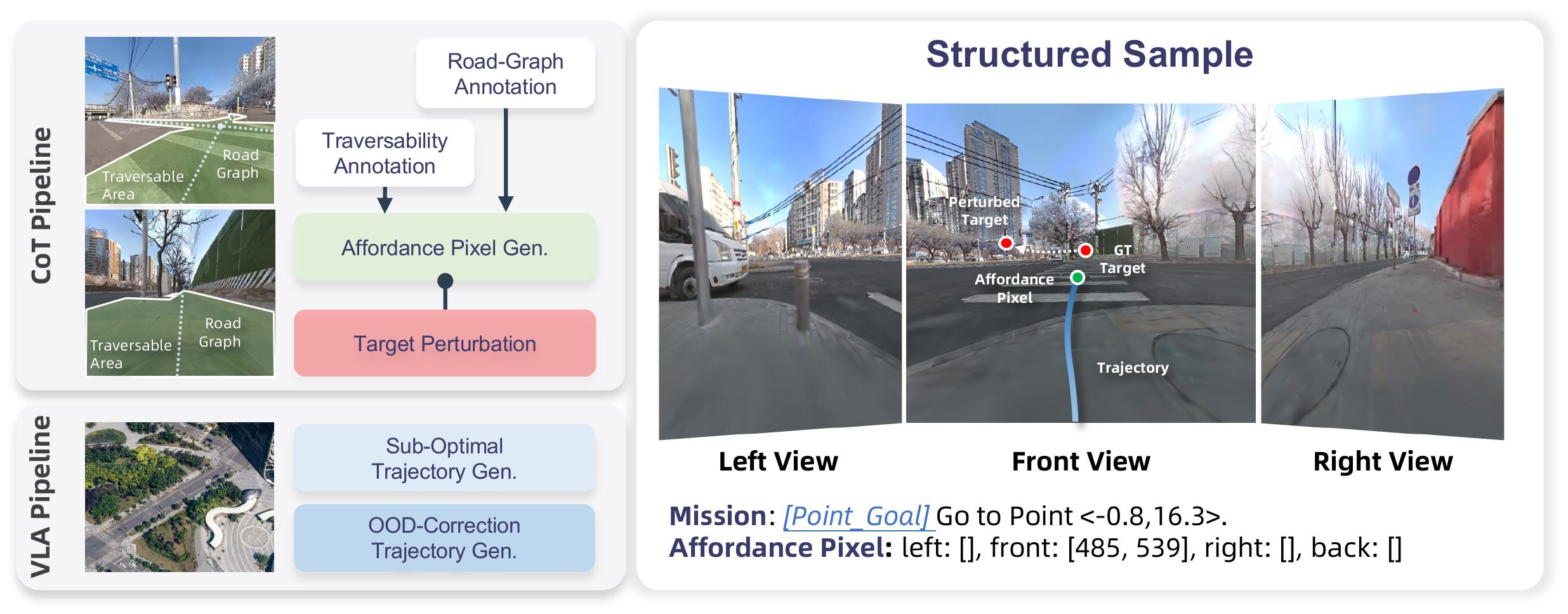}

\caption{\textbf{Data Construction Pipeline for the Point-Goal Corpus.}
Left: the data construction pipeline in two parts. The top half is the
CoT data construction, which generates affordance pixels from the
traversability and road-graph annotations and perturbs the target
coordinate; the bottom half is the VLN data construction, comprising
sub-optimal trajectory and OOD-correction trajectory synthesis. Right:
an example structured sample with tri-view observations and affordance
pixel annotation.}
\label{fig:point-data}
\end{figure*}

\textbf{Slow System ($\sim$2.36\,M).}
Its input consists of the tri-view observation, a noisy global
target, and---in 50\% of the samples---recent history frames paired
with the slow system's predictions on them as a temporal prior; the
remaining 50\% drops history to keep the model robust when memory is
absent.
Because  \textit{point-goal} targets are ego-centric offsets relative to
the robot's current pose, deviations arising from non-high-precision
map routes and localization inaccuracies can shift the apparent target
in the body frame---potentially placing it in a physically illegal
region (vehicle lane, flower bed, or other non-walkable area) rather
than on a valid pedestrian pathway.
To robustify the policy against this real-world coordinate drift, we
perturb the input target with a direction-aware non-uniform
distribution (normal component up to 10\,m, tangential up to 2\,m)
and reject perturbations whose angular deviation from the ground
truth exceeds $30^\circ$ to prevent semantic ambiguity. The model
learns to fall back on visual evidence and walkable-region semantics
to locate a legally reachable arrival point
whenever the input coordinate has drifted outside the traversable area.

\textbf{Fast System ($\sim$6.20\,M).}
Its input comprises the current tri-view frame, the current history,
the slow system's most recent affordance result together with its
reference views, and the global position target. The history channel
compensates for the asynchronous inference gap: because the slow
system runs at a lower rate than the fast system, recent reference
frames fill the information vacuum between slow updates and damp
control jitter. The supervision targets are the next 5 ego-centric
relative waypoints, with adaptive step sizes that are larger outdoors
and smaller indoors, trading off open-space throughput against
fine-grained obstacle avoidance. These waypoint supervisions are
drawn from both sub-optimal and OOD-correction trajectories, exposing
the controller to off-optimal states and teaching it to steer back
onto a valid path.
Figure~\ref{fig:point-data} illustrates the end-to-end data construction pipeline for the  \textit{point-goal} corpus.

\paragraph{Instruction-Following.}
We assemble the \textit{instruction-following} corpus from four complementary
streams curated in ABot-N0~\cite{abotn0}, all rendered in
Habitat~\cite{savva2019habitat} and our proprietary simulator. The dominant stream is
\textit{VLN-CE R2R/RxR}~\cite{krantz2020vlnce,ku2020rxr} on Matterport3D;
three auxiliary streams over the InteriorGS scene bank inject
short-horizon and grounding skills. Pixel supervision is
constructed by a unified geometric pipeline shared across all four
streams: along each reference trajectory, the affordance pixel is
the projection of the next traversable waypoint at a fixed
look-ahead horizon, and the target pixel is the projection of the
route endpoint, emitted only once the agent enters the final
approach segment to signal task termination. A VLM then validates each projected pixel, filtering out those that are not visible in the current view.

\begin{figure}[!t]
\centering
\includegraphics[width=\linewidth]{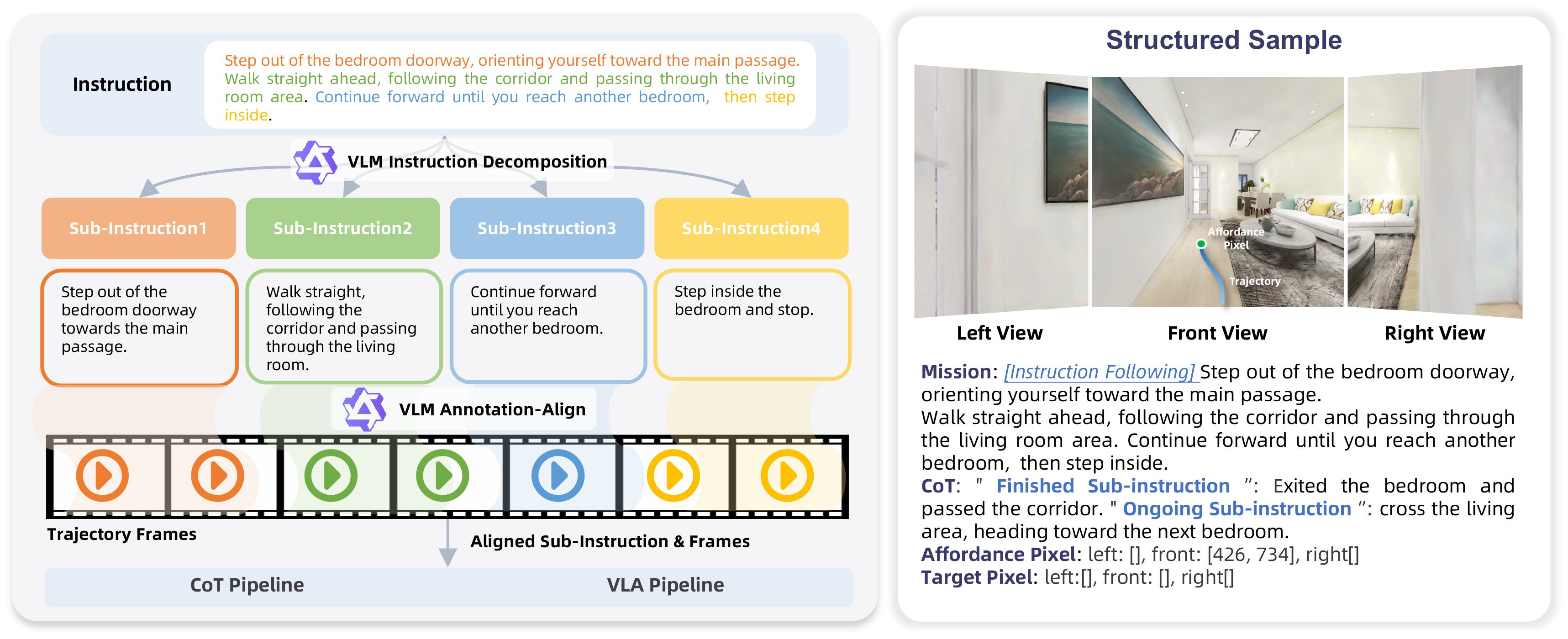}

\caption{\textbf{Data Construction Pipeline for the Instruction-Following Corpus.}
Left: a three-stage pipeline that decomposes long natural-language instructions into short sub-instructions, aligns each sub-instruction to its corresponding frame range along the milestone path, and generates and verifies affordance and target pixels for CoT and VLN data. Right: an example structured sample showing tri-view observations with the language instruction and pixel-level annotations for affordance and target.}
\label{fig:ins-data}
\end{figure}

\textbf{VLN-CE R2R/RxR (slow $\sim$2.25\,M, fast $\sim$2.10\,M).}
We build the corpus from $\sim$30\,K reference trajectories. CoT
labels are produced by a two-stage Qwen-3.5-27B annotator: Stage~1
decomposes each long natural-language instruction into an ordered
list of executable sub-instructions; Stage~2 aligns every
sub-instruction to its corresponding trajectory frame range,
yielding per-frame ``done / current'' progress annotations.
Pixel labels follow the unified geometric pipeline above and are
therefore decoupled from the instruction annotator (Figure~\ref{fig:ins-data}).

\textbf{InteriorGS Auxiliaries (slow $\sim$2.86\,M, fast $\sim$2.57\,M).}
Three InteriorGS-rendered streams sharpen specific competencies,
each annotated for both systems via the unified pixel pipeline:
(i) door-traversal clips (slow $\sim$0.44\,M, fast $\sim$0.38\,M)
for narrow-passage maneuvers; (ii) an office-building
\textit{instruction-following} set (slow $\sim$69\,K, fast $\sim$68\,K)
purpose-built for commercial-tower interiors (corridors, lobbies,
meeting rooms, and elevator halls), a domain that is severely
underrepresented in residential-centric R2R/RxR but critical for
real deployments such as last-mile indoor delivery;
(iii) short-horizon primitives (slow $\sim$2.35\,M, fast
$\sim$2.12\,M) covering atomic rotational, translational, and
compound actions to regularize fine-grained execution. The slow
count is consistently slightly larger than the fast count because
the slow corpus applies stronger terminal-segment
augmentation, where end-of-instruction frames are oversampled and
perturbed more aggressively so that the slow system reliably
recognizes the correct stopping point and emits the target pixel
at the right moment.

\textbf{CoT Supervision.}
The slow system's CoT reasons about instruction progress---what
sub-instruction has just been completed and what is currently being
executed---following the schema established in
ABot-N0. Combined with the unified affordance and target pixels,
this cleanly separates long-horizon path tracking from terminal
grounding.

\begin{figure*}[!t]
\centering
\includegraphics[width=\linewidth]{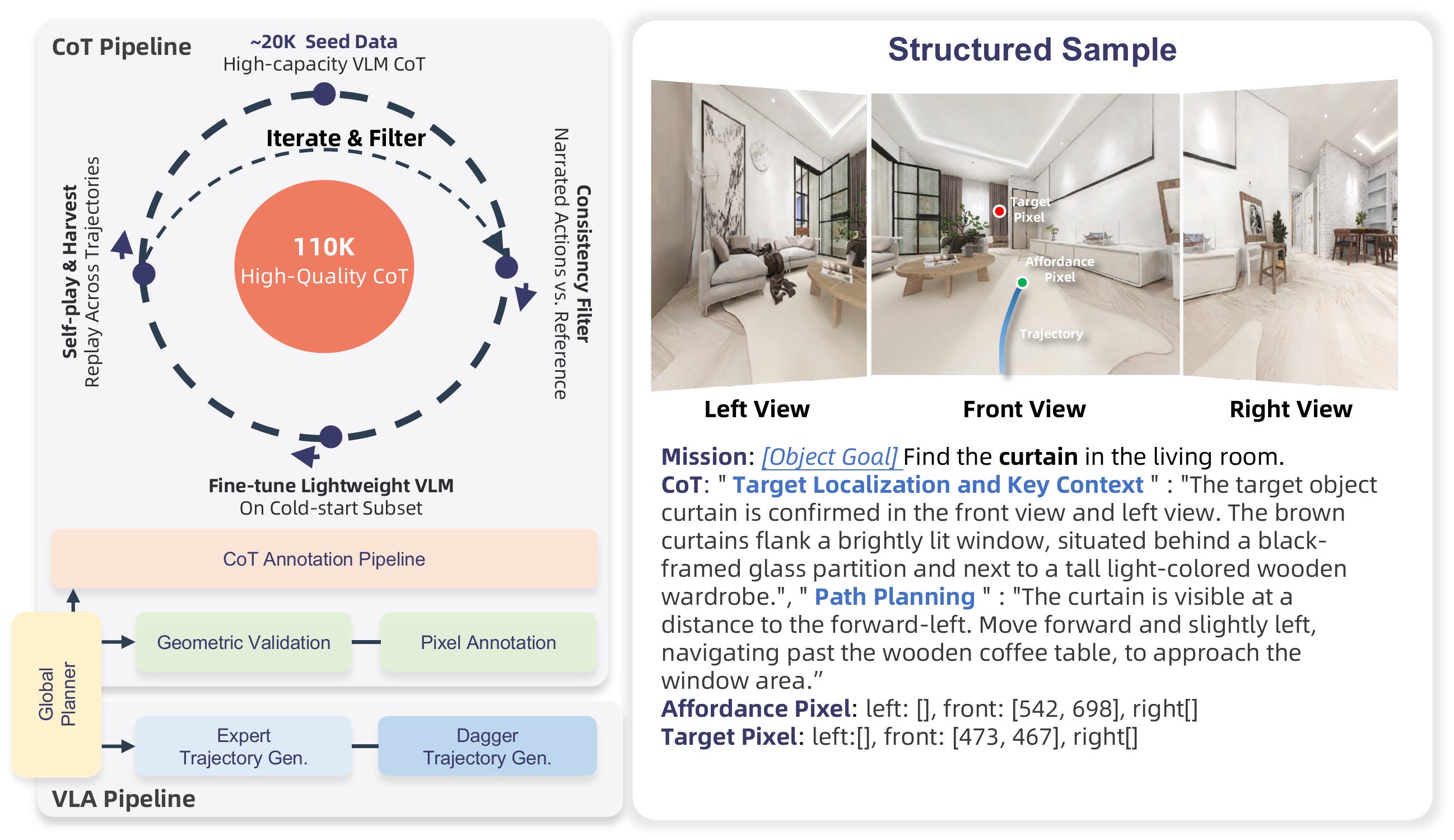}

\caption{\textbf{Data Construction Pipeline for the Object-Goal Corpus.}
The left panel comprises two parts: the top half illustrates the
iterative data flywheel that constructs the CoT rationales, scaling
high-capacity VLM data seeds to 110\,K high-quality structured samples
through A$^{\!*}$ consistency filtering and self-play harvesting; the
bottom half depicts the VLN pipeline that produces the low-level
supervision, including pixel annotation and OOD-correction trajectory
generation. The right panel demonstrates the resulting structured
tuple, featuring tri-view observations, explicit object and affordance
pixel grounding, and detailed two-block CoT rationales.}
\label{fig:object-data-flow}
\end{figure*}

\paragraph{Object-Goal.}
We construct the \textit{object-goal} corpus on the
OVON~\cite{yokoyama2024hmovon} 145-scene
Habitat~\cite{savva2019habitat} subset, complemented by proprietary
InteriorGS-rendered scenes curated in ABot-N0~\cite{abotn0} for
long-tail category coverage. Reference trajectories are produced
by A$^{\!*}$ from each starting pose to the target object, and
because the dual-system decomposition explicitly factors recognition
and approach apart, we keep the slow- and fast-system corpora
structurally distinct. Figure~\ref{fig:object-data-flow}
illustrates the overall data generation pipeline.

\textbf{Slow System ($\sim$110\,K).}
The slow system maps a tri-view observation plus an object name to
three co-emitted outputs: a natural-language path plan, an affordance
pixel, and an object pixel. The affordance pixel is the projection
of the next A$^{\!*}$ corner waypoint along the current segment onto
the camera image plane; the object pixel is the projection of the
target object's 3D centroid, emitted only once the object is
visible. CoT labels are produced through a data flywheel rather
than a one-shot annotator dump: a high-capacity VLM first seeds
$\sim$20\,K raw CoT samples spanning target visibility, spatial
location with explicit landmark references, path narration, and the
implied atomic actions; every seed is then scored by the consistency
between its narrated atomic-action sequence and the A$^{\!*}$
reference, and only high-consistency samples enter the cold-start
set. We then launch a \textit{data flywheel}: the cold-start set
is used to fine-tune a lightweight open-source VLM (Qwen-2.5-VL)
rather than the slow system itself; this distilled annotator is
replayed across the full trajectory pool to harvest fresh CoT at
scale, the outputs are re-scored against A$^{\!*}$ consistency,
and the filtered results are fed back to refine the annotator in
the next iteration. The loop converges at $\sim$110\,K
high-quality episodes, attenuating the seeding-VLM's
hallucinations while entirely removing the dependency on the
expensive high-capacity VLM at production time.

\textbf{Fast System ($\sim$2.10\,M).}
Inputs pair the current frame with a slow-system reference frame
(either the current frame or a recent history frame) carrying the
slow system's natural-language path plan, affordance pixel, and
object pixel. To prevent the fast controller from over-relying on
any single upstream signal, we randomly mask each of the three
reference channels independently during training, so the controller
learns to remain effective under partial slow-system guidance and
degrades gracefully when individual signals are missing or stale.
Outputs are 5 ego-centric relative waypoints at a fixed 0.3\,m/frame
step size, supervised by both sub-optimal and OOD-correction
trajectories so that the controller learns to recover from off-optimal
poses and out-of-distribution deviations.

\textbf{CoT Structure.}
The CoT decomposes into two named blocks aligned with the
dual-pixel output: \textit{Target Localization and Key Context}
describes whether the target is currently visible and, if so, its
spatial location relative to surrounding objects and the agent's
pose; \textit{Path Planning} reasons about the next actions required
to reach the target. The affordance pixel is always emitted to
supervise locomotion; once the target enters the field of view, the
\textit{object pixel} is emitted alongside it, providing explicit
visual grounding for the approach phase.

\begin{figure*}[!t]
\centering
\includegraphics[width=\linewidth]{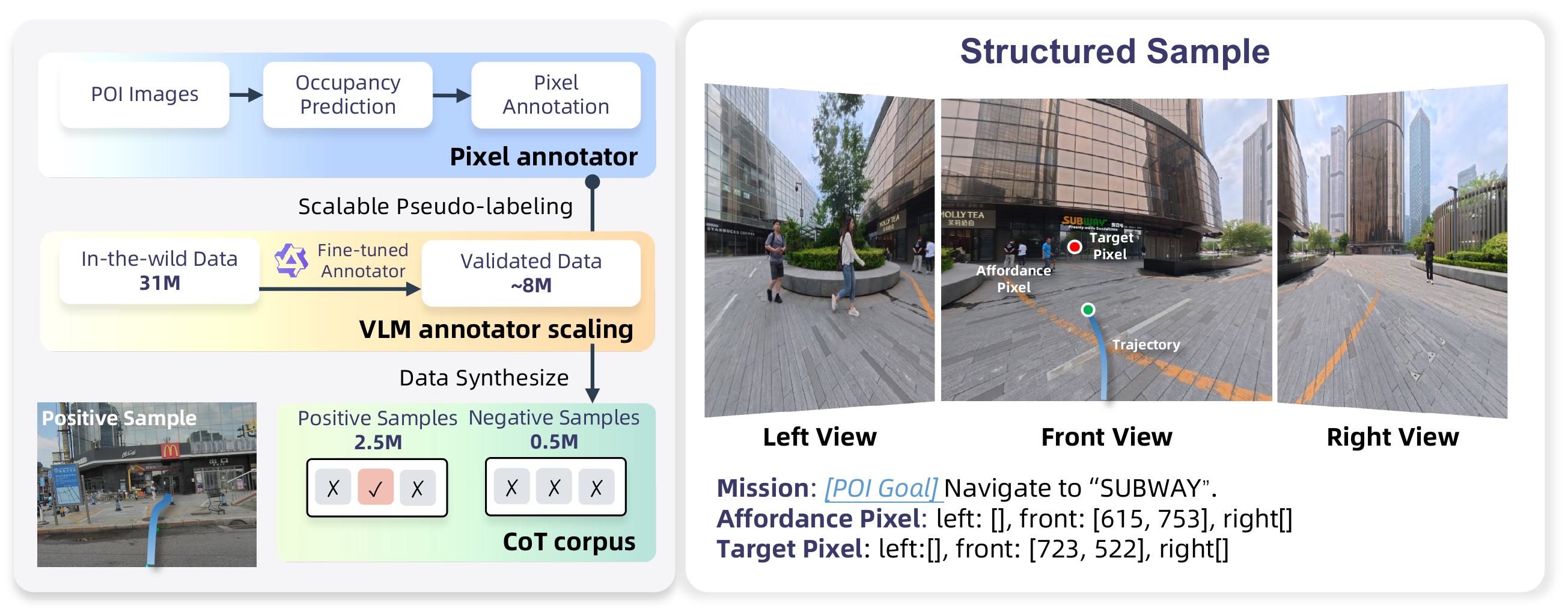}

\caption{\textbf{The Data Construction Pipeline for the POI-Goal Corpus}.
Left: the three-stage construction flow---generating geometric seed
annotations via monocular depth (Stage~1), scaling and filtering
31\,M street-view pairs using a distilled VLM (Qwen-3.5-4B) to yield
8\,M valid paths (Stage~2), and synthesizing tri-view episodes into
positive and negative sample pairs that harden the system's rejection
capability under missing-target conditions (Stage~3). Right: an
example structured sample.}
\label{fig:poi-pipeline}
\end{figure*}

\paragraph{POI-Goal.}
No 3D simulator currently covers the commercial-street settings
required for  \textit{POI-goal} navigation, so closed-loop rollout collection
is infeasible. We therefore supervise only the slow system on static
street-view annotations and let the fast system inherit its execution
capability from \textit{point-goal} at inference time. To reach
data-engine scale, we build a three-stage annotation pipeline
(Figure~\ref{fig:poi-pipeline}).

\textbf{Seed Annotation ($\sim$275\,K).}
We collect tuples of (POI name, signage bbox, entrance bbox) from
two sources: 55\,K human-annotated samples from BridgeNav with
first-frame signage and entrance bounding boxes, and 220\,K samples
from POINav with rule-plus-model pseudo-labels covering roughly
70\,K distinct images (each image may host multiple POIs).
We run MoGe-V2 \cite{wang2025moge} monocular depth on every image, derive an
image-aligned occupancy map from surface normals, set the target
pixel to the nearest ground point under the entrance-bbox bottom
center, and trace a straight ground-line from the camera footprint
to the target, sampling waypoints every 0.5\,m and truncating at
the first obstacle.

\textbf{VLM Annotator Scaling.}
The MoGe-V2 pipeline is too slow to scale to tens of millions of
images, so we distill it into a Qwen-3.5-4B annotator (full-LLM
fine-tuning with the ViT and projector frozen). The annotator takes
a POI name and a target affordance distance bucket
($\{2,3,4,5\}$\,m) and emits a JSON record whose ordered fields
(signage bbox $\rightarrow$ entrance bbox $\rightarrow$ target pixel
$\rightarrow$ affordance pixel) act as an implicit chain of thought.
Training covers three patterns: positives where a feasible
affordance pixel exists, positives where the path is blocked
(no feasible affordance pixel), and negatives where the queried POI
is absent from the image, instilling an explicit
``cannot-be-grounded'' rejection behavior. Applying this annotator
to $\sim$31\,M collected street-view (POI name, image) pairs
at the 3\,m bucket yields $\sim$8\,M valid positive labels.

\textbf{N1 Slow-System Corpus ($\sim$3.0\,M).}
Because N1 consumes a tri-view input but the upstream pipeline is
monocular, we synthesize tri-views by stitching one positive image
with two negative images (the queried POI absent from the two side
views). The final slow-system corpus contains 2.5\,M positive
tri-view episodes and 0.5\,M fully negative tri-view episodes (POI
absent in all three views) to harden rejection capability under
missing-target conditions. \textit{POI-goal} is a single-shot grounding task
that does not benefit from temporal memory, so the slow system
operates on the current observation only and history frames are
omitted.

\begin{figure}[!t]
\centering
\includegraphics[width=\linewidth]{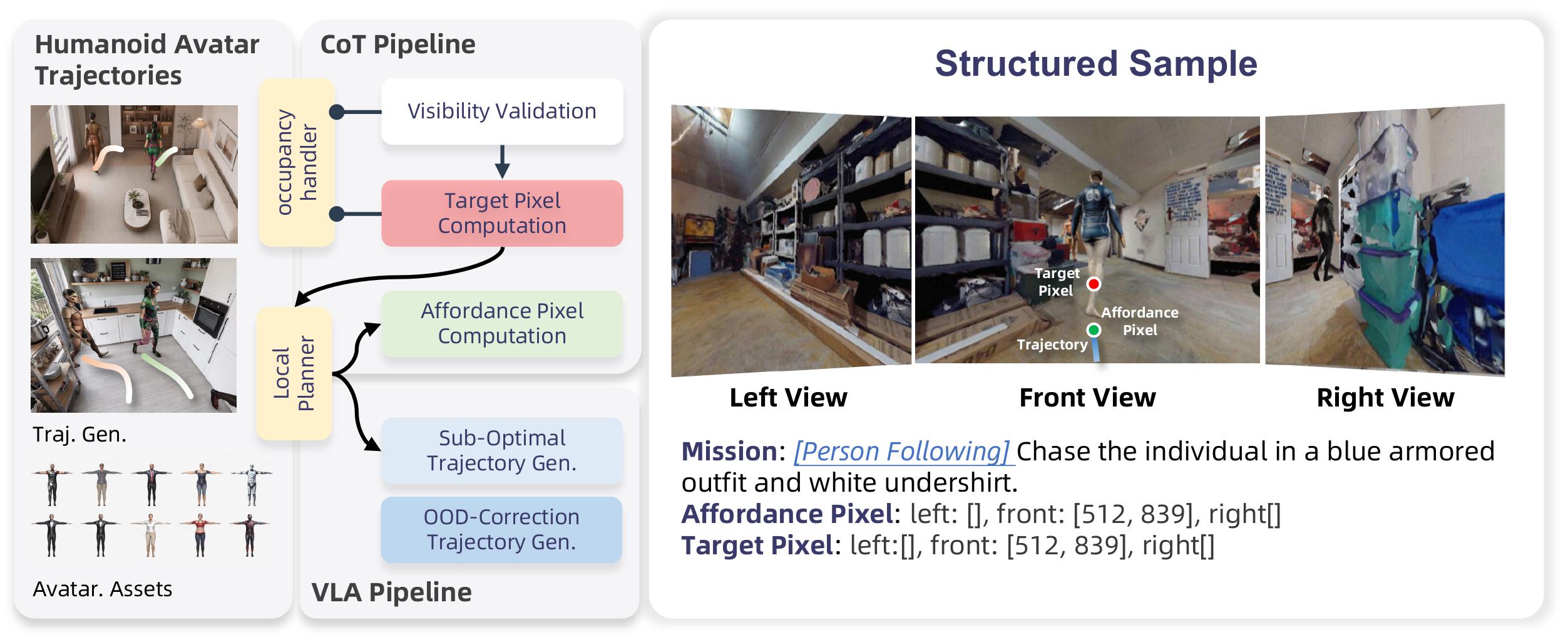}

\caption{\textbf{Data Construction Pipeline for the Person-Following Corpus.}
Left: the data construction pipeline covering both CoT and VLN data.
The pixel (CoT) data derives affordance and target pixels from human
avatar trajectories through A$^*$ waypoint planning, visibility
detection, and stochastic prediction perturbation, while the VLN data
comprises sub-optimal trajectory and OOD-correction trajectory
synthesis. Right: an example structured sample showing tri-view
observations with affordance and target pixel annotations, along with
the language instruction specifying the target appearance.}
\label{fig:person-data}
\end{figure}

\paragraph{Person-Following.}
Following TrackVLA~\cite{wang2025trackvla}, we generate
avatar-tracking sequences in Habitat~\cite{savva2019habitat}
photorealistic indoor scenes by replaying TrackVLA's open-sourced
humanoid motion trajectories and planning agent waypoints with
A$^{\!*}$ at three following distances (1.2, 1.5, and 2.0\,m). Specific data generation procedures are illustrated in Figure~\ref{fig:person-data}.

\textbf{Scenario Composition.}
For each following distance we synthesize three scenario categories
that mirror TrackVLA's taxonomy:
\textit{Single-Target Tracking} (STT) with a unique target,
\textit{Distracted Tracking} (DT) with non-target distractor avatars,
and \textit{Ambiguity Tracking} (AT) where multiple avatars share
attributes with the target so that disambiguation must be resolved
from the language query.

\textbf{Slow System ($\sim$2.7\,M).}
We draw samples from the 1.2\,m and 2.0\,m sets. When the target
is visible, both the affordance and target pixels are anchored to
the bottom-center of the target bounding box; once the target leaves
the field of view, the target pixel is set to null and the
affordance pixel falls back to the first projectable future
waypoint, forcing the model to reason about occluded targets
through the memory buffer.

\textbf{Fast System ($\sim$3.4\,M).}
We span all three distances and randomly pair the current frame
with a reference frame sampled $0$--$4$ steps earlier so that the
fast system learns to consume slightly stale slow-system
predictions. In addition to the waypoint sequence the fast system
also predicts per-step heading and a safety-clearance flag. The
waypoint supervision is likewise built from sub-optimal and
OOD-correction trajectories, training the controller to recover
smoothly when the followed target drifts off the optimal path.

\textbf{Asymmetric Noise Schedules.}
Both systems are trained with task-specific perturbations over the
prior pixel inputs. The slow system tolerates an aggressive
schedule because its temporal context smooths out corrupted single
frames; the fast system, far more sensitive to single-frame pixel
perturbations, uses a conservative schedule. This asymmetry keeps
both policies stable under imperfect upstream signals and
transient detection failures at deployment.

\subsection{Post-Training}
\label{sec:method-posttraining}

Supervised pre-training aligns the slow system's outputs with expert
demonstrations, yet these demonstrations are inherently limited:
reference trajectories produced by shortest-path planners often lack
safety margins and ignore implicit social constraints. More
fundamentally, token-level supervision does not directly optimize the
downstream navigation outcome, leaving a gap between imitating
individual decisions and maximizing episode-level success. To bridge
this gap, we apply Group Relative Policy Optimization
(GRPO)~\cite{deepseekai2025r1} to the slow system, casting its
pixel-goal generation as a language-model policy whose outputs are
rewarded by environment-grounded signals.

\subsubsection{Formulation}
\label{sec:grpo-formulation}

For each state $s=(I^{\mathrm{mem}}, I^{\mathrm{cur}}_{\mathrm{tri}}, g)$
we sample $G$ candidates $o^{(i)}{=}(\mathcal{C}^{(i)},\mathbf{p}^{(i)})\sim\pi_\theta$, score each with the composite reward $R^{(i)}$ defined
below, and form group-relative advantages
$A^{(i)}{=}(R^{(i)}{-}\mu_R)/(\sigma_R{+}\varepsilon)$ over the group.
We then optimize
\begin{equation}
\mathcal{L}_{\mathrm{GRPO}}(\theta) \;=\;
\mathbb{E}\!\left[\sum_{i,t}
\min\!\bigl(\rho_t^{(i)} A^{(i)},\;
\mathrm{clip}(\rho_t^{(i)},\,1{\pm}\epsilon)\,A^{(i)}\bigr)
\right]
\;-\;\beta\,\mathrm{KL}\!\bigl[\pi_\theta\,\Vert\,\pi_{\mathrm{ref}}\bigr],
\label{eq:grpo}
\end{equation}
with $\rho_t^{(i)}$ the per-token importance ratio against the previous
policy iterate. The KL anchor against the reference model
$\pi_{\mathrm{ref}}$ is non-negotiable in our setting: aggressive
exploration in the joint CoT$+$pixel output space quickly produces
unparseable strings, on which every downstream reward becomes zero
and the gradient collapses.

\subsubsection{Reward Design}
\label{sec:grpo-reward}

The composite reward
$R\!=\!w_f R_{\mathrm{format}}\!+\!w_t R_{\mathrm{target}}\!+\!w_o R_{\mathrm{safety}}$
(weights summing to one) targets three orthogonal failure modes of
the pretrained policy and is calibrated to a comparable scale so that
within-group normalization remains informative.

\paragraph{Format.}
A binary indicator that the output respects the per-view JSON schema.
We gate $R_{\mathrm{target}}$ and $R_{\mathrm{safety}}$ on
$R_{\mathrm{format}}{=}1$, preventing the policy from drifting off
the parseable manifold during early-stage exploration.

\paragraph{Target Alignment.}
The target reward is computed as the average of per-view matching scores. Missing predictions in any view are treated as escape behaviors and incur a large penalty. When valid points are present across all views, the reward adopts an exponential decay of their L2 distance. The overall formulation is given by:
\begin{equation}
    R_{\mathrm{target}}\!=\!\sum_{\mathrm{frame}}\alpha_t\ {\exp\bigl(-\|\hat{\mathbf{p}}-\mathbf{p}^{\star}\|^2/\mathrm{scale}\bigr)}_{\mathrm{frame}}, \quad \mathrm{frame}\in\{\mathrm{front}, \mathrm{left}, \mathrm{right}\},
\end{equation}
where $\hat{\mathbf{p}},\mathbf{p}^{\star} \in \mathbb{R}^2$ denote the predicted and ground-truth target coordinates in each camera view, respectively; $\mathrm{scale}$ controls the spatial tolerance of the exponential kernel, determining how rapidly the reward decays with increasing prediction error; and $\alpha$ is a normalization weight that ensures balanced contributions across views when computing the final averaged reward. This design effectively constrains model outputs and mitigates knowledge forgetting during  pretraining.

\paragraph{Safety Clearance.}
We back-project $\hat{\mathbf{p}}$ to the world frame via the local
depth map and camera intrinsics, take the worst-case clearance $d$
across views, and apply
\begin{equation}
R_{\mathrm{safety}}(d) =
\begin{cases}
-\alpha_0, & d \ge d_{\mathrm{safe}},\\
-\alpha_o\,\exp\!\bigl(\beta\,(d_{\mathrm{safe}}{-}d)\bigr), & d < d_{\mathrm{safe}}.
\end{cases}
\label{eq:obstacle}
\end{equation}
Here $d$ measures the distance to the nearest non-traversable region,
which encompasses not only physical obstacles but also traffic-rule
violations and illegal zones (e.g.\ vehicle lanes, restricted
construction areas). The exponential profile is deliberately
asymmetric: flat in safe zones, so legitimate close-quarters
maneuvers (e.g.\ doorway traversal) are not penalized, but sharp
inside the danger margin where behavioral correction is most urgent.
Catastrophic outcomes (parse failure, full intrusion into prohibited
regions) are clamped to a large negative floor so that they always
receive the worst within-group advantage.

\subsubsection{Balance Sampling for GRPO Training Dataset}
\label{sec:grpo-trainingdata}

As shown in the post-training composition of Figure~\ref{fig:data}, we
build the GRPO training set on top of the  \textit{point-goal} pretraining
corpus: episodes are stratified by proximity to non-traversable
regions into Safe, Critical, and Danger zones, unstable samples beyond
the stability limit are discarded, and the remaining episodes are
sampled according to a theoretically motivated ratio to ensure
balanced gradient signals.

To formalize this balancing strategy, we analyze the geometric properties of the sample distribution by examining the local variance of the composite reward. For a given prompt, the policy generates $K$ rollout trajectories within a group $G$; let $d$ denote the distance from the robot to the closest non-traversable region (including physical obstacles, illegal zones, and traffic-rule-violating areas) along each rollout. Due to policy stochasticity, these $K$ rollouts yield different trajectories and thus different $d$ values, with within-group mean $\bar{d}$ and variance $\sigma_d^2$. When the group mean penetrates the safety margin ($\bar{d} < d_{\mathrm{safe}}$), we apply a first-order Taylor expansion to the asymmetric exponential penalty $R_{\mathrm{safety}}(d) = -\alpha_o \exp\bigl(\beta(d_{\mathrm{safe}}-d)\bigr)$ around $\bar{d}$:
\begin{equation}
    R_{\mathrm{safety}}(d) \approx R_{\mathrm{safety}}(\bar{d}) + \alpha_o \beta \exp\bigl(\beta(d_{\mathrm{safe}}-\bar{d})\bigr) (d - \bar{d}).
\end{equation}

Since the model output remains valid in non-degenerate cases, we assume $\mathrm{Var}(R_{\mathrm{format}}) \approx 0$. Taking the non-zero covariance between the target alignment and safety clearance into account, the total within-group variance $\sigma_G^2$ (omitting the early-converged format term) is formulated as:
\begin{equation}
    \sigma_G^2 \approx w_t^2 \mathrm{Var}(R_{\mathrm{target}}) + w_o^2 \mathrm{Var}(R_{\mathrm{safety}}) + 2w_t w_o \mathrm{Cov}(R_{\mathrm{target}}, R_{\mathrm{safety}}),
    \label{eq:total_var}
\end{equation}
where the local variance propagated by the safety term is approximated by:
\begin{equation}
    \mathrm{Var}(R_{\mathrm{safety}}) \approx \left[ \alpha_o \beta \exp\bigl(\beta(d_{\mathrm{safe}}-\bar{d})\bigr) \right]^2 \sigma_d^2.
    \label{eq:obstacle_var}
\end{equation}

Due to the intrinsic task conflict in close-quarters maneuvers, these two objectives exhibit a strong negative correlation, i.e., $\mathrm{Cov}(R_{\mathrm{target}}, R_{\mathrm{safety}}) = -\rho \sigma_{\mathrm{target}}\sigma_{\mathrm{safety}}$ with $\rho > 0$. Although this negative covariance slightly dampens the aggregate variance locally, the squared derivative in Eq.~\eqref{eq:obstacle_var} still dictates that the safety reward variance grows exponentially with the intrusion depth $(d_{\mathrm{safe}}-\bar{d})$ at rate $2\beta$, even when $\sigma_d^2$ itself is small for deeply intruding prompts.

To prevent this exploding variance from masking the target alignment gradients, the gradient isotropy condition dictates that the variance contribution of the target alignment reward must be bounded by a stability threshold $\gamma \in (0, 1)$, such that:
\begin{equation}
    w_t^2 \mathrm{Var}(R_{\mathrm{target}}) \geq \gamma\ \sigma_G^2.
\end{equation}

Substituting Eq.~\eqref{eq:total_var} into the current equation yields:
\begin{equation}
    w_o^2 \mathrm{Var}(R_{\mathrm{safety}}) - 2w_t w_o \rho \sigma_{\mathrm{target}}\sigma_{\mathrm{safety}} \le \frac{1-\gamma}{\gamma} w_t^2 \mathrm{Var}(R_{\mathrm{target}}).
\end{equation}

By solving this quadratic inequality with respect to $\sigma_{\mathrm{safety}}$, we can map out the minimum allowable average distance $\bar{d}_{\min}$ that defines the critical hazard boundary for a valid prompt. Prompts with $d \le \bar{d}_{\min}$ produce reward variance that exceeds the stability budget and are therefore discarded from the training set. Introducing a transition margin $\Delta d > 0$, we partition the remaining prompts into three categories:
\textit{Safe Zone} ($d \ge d_{\mathrm{safe}}$): the exponential penalty is inactive, contributing negligible reward variance;
\textit{Critical Zone} ($\bar{d}_{\min} + \Delta d < d < d_{\mathrm{safe}}$): the penalty is active but the resulting variance remains well within the stability budget;
\textit{Danger Zone} ($\bar{d}_{\min} < d \le \bar{d}_{\min} + \Delta d$): the variance approaches the tolerable limit, yet the steep penalty slope produces the strongest directional gradient signal among all trainable prompts.


To curate a well-balanced training set, these scenarios must be mixed in a controlled ratio guided by the Gradient Signal-to-Noise Ratio (GSNR). In GRPO, training stability requires not only bounding the aggregate variance but also ensuring sufficient gradient magnitude across all critical regions. 
These three zones exhibit fundamentally distinct gradient characteristics. Safe Zone prompts serve as high-GSNR anchors that stabilize the baseline target alignment gradient. Critical Zone prompts suffer from degraded GSNR due to exponential variance growth and negative target-safety covariance, yet they provide essential boundary refinement signals. Crucially, Danger Zone prompts produce the largest gradient magnitude among all trainable samples via the steep exponential penalty slope; although their per-sample GSNR is the lowest (variance grows faster than signal at rate $2\beta$ vs.\ $\beta$), batch averaging over $P_{\mathrm{danger}} \times B$ samples allows the consistent repulsive direction to accumulate while noise partially cancels. This directional signal is irreplaceable: without it, the policy risks converging to suboptimal solutions near the safety boundary, as the milder gradients from the Critical Zone alone are insufficient to enforce hard safety constraints.
Therefore, the sampling proportions must satisfy a composite gradient sufficiency condition rather than a simple variance equality:
\begin{equation}
    P_{\mathrm{safe}} \cdot \mathcal{S}_{\mathrm{safe}} + P_{\mathrm{critical}} \cdot \mathcal{S}_{\mathrm{critical}} + P_{\mathrm{danger}} \cdot \mathcal{S}_{\mathrm{danger}} \geq \tau,
    \label{eq:composite_gradient}
\end{equation}
where $\mathcal{S}_z = |\mu_{\nabla,z}|^2 / \sigma_{\nabla,z}^2$ denotes the per-sample GSNR of zone $z$ (expected gradient magnitude squared divided by gradient variance), and $\tau$ is a minimum threshold for stable policy improvement. 

We summarize the principles guiding our sampling proportions as follows:
\begin{itemize}
    \item Safe Zones take up the largest share. They stabilize training by providing high‐GSNR signals that prevent model degradation.
    \item Critical Zones are kept in check to avoid inflating batch‐variance with their moderate‐GSNR samples, yet—together with Danger Zones—they supply essential safety-compliance rewards.  
    \item Danger Zones, although they exhibit the lowest per-sample GSNR, deliver a unique and irreplaceable gradient direction for hard-safety enforcement. To preserve overall training stability, they are assigned the smallest proportion.
\end{itemize} 
Empirically, we adopt a structured sampling ratio of $5:3:2$ ($P_{\mathrm{safe}} : P_{\mathrm{critical}} : P_{\mathrm{danger}}$). This ratio ensures that: (1) $P_{\mathrm{safe}} = 0.5$ provides dominant gradient stability; (2) $P_{\mathrm{critical}} = 0.3$ supplies adequate boundary learning without exceeding the variance budget; and (3) $P_{\mathrm{danger}} = 0.2$ delivers sufficient high-magnitude repulsive gradients to prevent unsafe convergence. This theoretically grounded composition enables stable GRPO training by balancing gradient fidelity, variance control, and safety enforcement simultaneously.

\subsubsection{Training Dynamics and Generality}
\label{sec:grpo-data}

To prevent degradation of the capabilities acquired during the pretraining phase, we avoid biased preference sampling in GRPO data collection. We employ action-balanced batch sampling to ensure dataset unbiasedness. By balancing trivial states with challenging scenarios (e.g., near non-traversable regions), this strategy not only preserves the pretrained policy's generalization but also smooths within-group learning signals, thereby stabilizing training. We address two reward-hacking equilibria: \textit{target withholding} (avoiding goals to minimize safety penalty) and \textit{permanent inaction} (zero motion and progress). To mitigate these, we explicitly balance the magnitudes of $R_{\mathrm{target}}$ and $R_{\mathrm{safety}}$ to prevent disproportionate avoidance behavior, and impose a high-magnitude missing-prediction penalty within $R_{\mathrm{target}}$. This combination effectively breaks both inertial and evasive stalemates.

The factorization is task-agnostic by construction: $R_{\mathrm{format}}$ depends only on the output schema and $R_{\mathrm{safety}}$ only on scene geometry and traversability rules, both shared across the five tasks. Extension to \textit{instruction-following}, \textit{object-goal}, or \textit{person-following} reduces to redefining $R_{\mathrm{target}}$ alone (route-endpoint distance, mask IoU, or bounding-box-center distance, respectively). We instantiate and validate the pipeline at scale on  \textit{point-goal} (0.5\,M episodes after zone classification and filtering, as shown in Figure~\ref{fig:data}) and view extension to the remaining tasks as a matter of metric design rather than algorithmic modification.

\section{Benchmark}
\label{sec:benchmark}

\begin{figure*}[t]
\centering
\includegraphics[width=\linewidth]{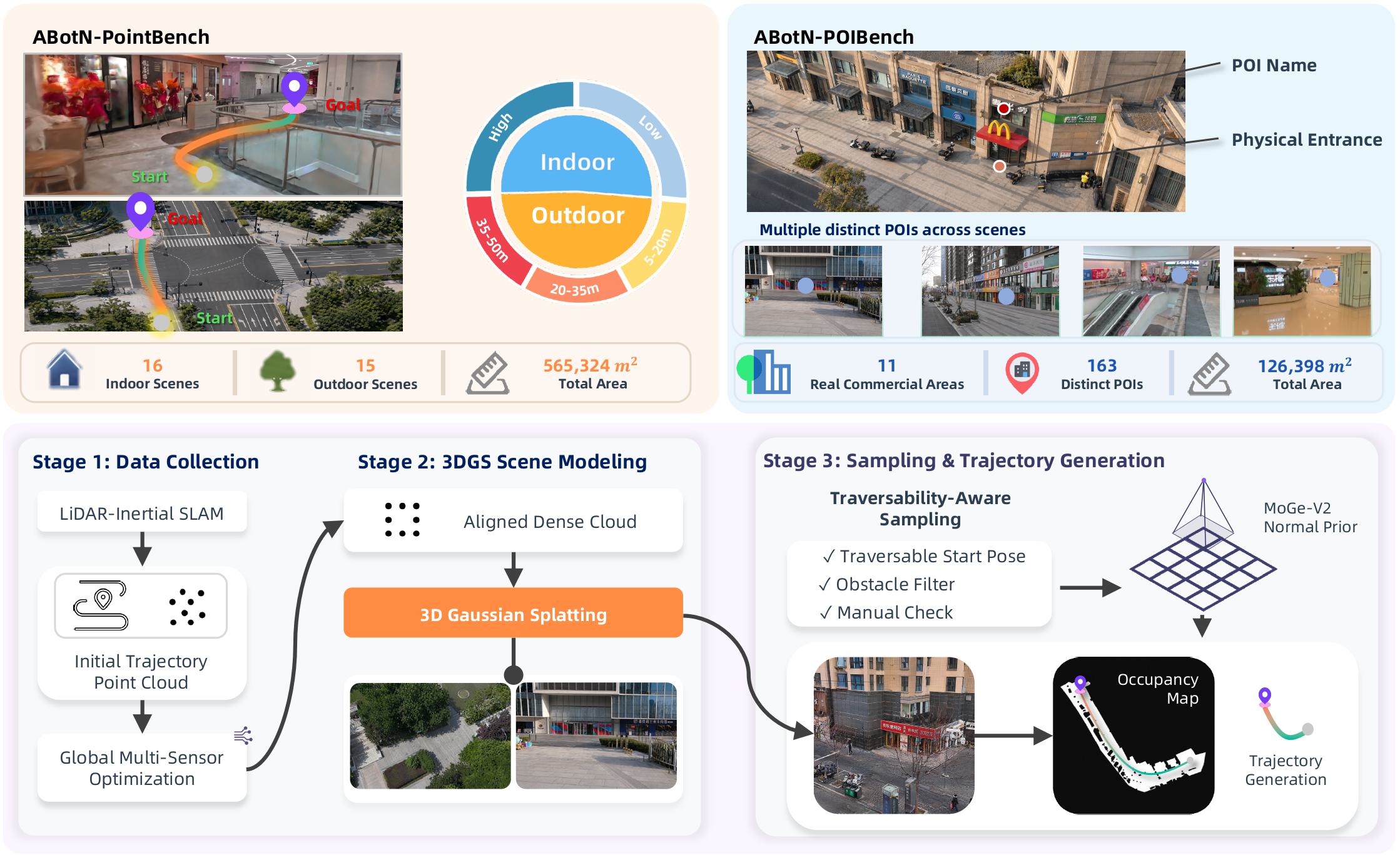}

\caption{\textbf{Overview of the ABotN Benchmark Suites and their
Unified Scene Construction Pipeline.} Top: Dataset
statistics and hierarchical distance splits for \textbf{ABotN-PointBench}
(left) and
\textbf{ABotN-POIBench} (right). Bottom: The unified three-stage generation
pipeline: (1) high-fidelity data collection via LiDAR-inertial
SLAM; (2) photorealistic 3DGS scene modeling initialized by
aligned dense point clouds; and (3) traversability-aware query
sampling and ground-truth reference trajectory generation using
A$^{\!*}$ on MoGe-V2-derived 2D occupancy grids.}
\label{fig:benchmarks}
\end{figure*}

Existing navigation benchmarks fall short of evaluating two
capabilities that are central to deployable real-world agents:
(i) closed-loop, social-rule-aware navigation in heterogeneous
indoor and outdoor environments, and
(ii) final-meters arrival at the physical entrance of a named
point of interest in dense commercial streetscapes. We therefore
introduce two complementary benchmarks built on the same
high-fidelity 3DGS reconstruction stack: \textbf{ABotN-PointBench}
for coordinate-conditioned navigation and \textbf{ABotN-POIBench} for
name-conditioned POI navigation. Figure~\ref{fig:benchmarks} gives
an overview of both benchmarks and the unified scene construction
pipeline shared by them. Then we describe the ABotN-PointBench and ABotN-POIBench in
detail, respectively.

\subsection{ABotN-PointBench}
\label{sec:bench-pointgoal}

\paragraph{Motivation.}
Most prior \textit{point-goal} benchmarks suffer from three structural
gaps. \emph{Scene homogeneity}: representative works such as
CityWalker evaluate exclusively on street blocks, leaving malls,
parks, and indoor commute settings unmeasured. \emph{Open-loop
protocol}: single-step waypoint regression decouples prediction
from downstream control and ignores compounding errors that
dominate real deployment. \emph{Social rules ignored}: traditional
benchmarks score only geometric arrival, so policies that cut
across vehicle lanes or trample lawns are rewarded equally with
socially compliant ones. ABotN-PointBench is the first
closed-loop, high-fidelity benchmark that explicitly addresses all
three gaps for indoor and outdoor \textit{point-goal} navigation.

\paragraph{Scene Construction.}
We collect 31 real-world scenes (16 indoor, 15 outdoor) spanning
shopping malls, parks, road intersections, and daily commute
corridors, and reconstruct each as a high-fidelity 3DGS
environment. Annotators then label every scene with a
fine-grained walkability map that distinguishes legally
traversable regions (sidewalks, crosswalks, building entrances,
intersection paths) from violating regions (vehicle and
non-motorized lanes, lawns, static obstacles such as bicycles,
pillars, and flower beds). The
resulting occupancy maps double as ground truth for both episode
generation and compliance scoring.

\paragraph{Episode and Reference-Trajectory Generation.}
For every scene we run A$^{\!*}$ on the annotated occupancy map to
produce shortest feasible paths used as reference trajectories for
efficiency metrics (e.g., SPL), yielding 465 distinct reference
trajectories.
Targets are stratified by distance
to cover both short-horizon precision and long-horizon endurance:
indoor targets are sampled in $3$--$20$\,m, while outdoor targets
are partitioned into three buckets ($5$--$20$, $20$--$35$,
$35$--$50$\,m). Each episode is rolled out closed-loop in our
in-house 3DGS evaluator, with the agent receiving real-time RGB
and a global coordinate.

\paragraph{Difficulty Stratification.}
To probe robustness under increasing navigational demands, episodes
are stratified by path length.
\emph{Outdoor} (15~scenes, 5 episodes per scene per tier):
Low ($5$--$20$\,m), Medium ($20$--$35$\,m), High ($35$--$50$\,m),
yielding 75 episodes per tier. Because the large-scale outdoor
occupancy annotations inevitably carry slight boundary deviations,
the outdoor split adopts a relaxed collision budget---an episode is
counted as successful only if the agent arrives with fewer than three
collision events; we denote this metric SR$_{<3\text{col}}$. This
redundant tolerance prevents minor labeling inaccuracies near
obstacle boundaries from being misinterpreted as genuine navigation
failures.
\emph{Indoor} (16~scenes, 8~Low-difficulty + 8~High-difficulty):
targets range over $3$--$20$\,m.  Additionally, since indoor scenes
are compact and precisely annotated, indoor episodes
impose a strict collision budget---an episode is counted as
successful only if the agent arrives with no collision
event; we denote this metric SR$_{<1\text{col}}$.

\paragraph{Metrics.}
We report the following metrics:
\begin{itemize}[nosep,leftmargin=1.2em]
\item \textbf{SR} (Success Rate): fraction of episodes where the
  agent reaches within a threshold distance of the target under the
  split-specific collision budget (SR$_{<3\text{col}}$ outdoors,
  SR$_{<1\text{col}}$ indoors).
\item \textbf{SPL} (Success weighted by Path Length): SR discounted
  by path efficiency relative to the A$^{\!*}$ reference trajectory.
\end{itemize}


\subsection{ABotN-POIBench}
\label{sec:bench-poigoal}

\paragraph{Motivation.}
\textit{POI-goal} navigation requires the agent to reach the
\emph{physical entrance} of a named POI, not merely an abstract
coordinate or a coarse semantic region. Existing protocols sit at
two unsatisfactory extremes: open-loop benchmarks like BridgeNav
evaluate only single-step waypoint prediction and cannot reflect
closed-loop interaction, while coarse-grained benchmarks like
CitySeeker evaluate only at the block or graph-node level and
miss the entrance-level final-meters arrival accuracy that real
deployment demands. ABotN-POIBench is, to our knowledge, the first
closed-loop, high-fidelity benchmark for real-world \textit{POI-goal}
navigation.

\paragraph{Scene Construction.}
We select 11 real commercial regions with high POI density and
diverse storefront types. Sidewalk width is treated as a hard
constraint during selection: robots have limited camera
heights and vertical FOV, so excessively narrow streets
would make high-mounted signage invisible from the agent's
perspective. Each region is reconstructed as a 3DGS scene through
a four-stage pipeline: (i) LiDAR-Inertial SLAM yields an initial
trajectory and point cloud; (ii) global multi-sensor optimization
jointly refines camera poses and produces a dense LiDAR point
cloud aligned with the captured RGB frames; (iii) the dense
LiDAR points directly initialize the 3D Gaussians, which are then
optimized against the RGB observations and refined camera poses
for improved geometric consistency, sharply improving
reconstruction quality on textureless street surfaces; (iv) a
dense triangle mesh is extracted from the optimized Gaussians and
registered with our in-house 3DGS evaluator as collision
geometry, ensuring both photorealistic fidelity and physically
faithful interaction.

\paragraph{POI and Entrance Annotation.}
The benchmark covers 11 commercial regions totaling
$\sim$126{,}398\,m$^2$, 163 distinct POIs, $>$38\,M Gaussian
points, $\sim$2.81\,M mesh vertices, and $\sim$4.40\,M mesh faces.
Each POI is annotated with both its name and its physical
entrance: the entrance is represented as a ground-flush horizontal
bounding frame defined by two ground-aligned points that span the
doorway width. This entrance frame serves as the ground-truth
arrival region against which final-meters success is judged,
avoiding the ambiguity of a single arrival coordinate.

\paragraph{Start Sampling and Reference Trajectories.}
For every annotated POI, we sample multiple starting poses within a
radius around the entrance under traversability-aware constraints:
(i) start point must lie inside human-annotated walkable regions;
(ii) grass, steps, and large obstacles unsuited to quadruped
locomotion are excluded; (iii) every start point is manually checked to
guarantee that, from the low viewpoint, at least
partial signage of the target POI is visible. To enable efficiency
metrics, we generate ground-truth reference trajectories by
estimating surface normals with MoGe-V2, classifying upward-facing
surfaces as traversable ground and walls/pillars as obstacles,
projecting to a 2D top-down occupancy grid, and running A$^{\!*}$
from the start to the entrance frame.

\paragraph{Metrics.}
We report success rate at a $2$\,m entrance threshold (SR$_{<2\mathrm{m}}$)
and SPL against the A$^{\!*}$ reference. The strict
entrance-distance criterion enforces final-meters arrival accuracy
and prevents trivial credit for stopping in the rough vicinity of
the storefront.

\section{Experiments}
\label{sec:experiments}

\setlength{\floatsep}{8pt plus 2pt minus 2pt}
\setlength{\textfloatsep}{10pt plus 2pt minus 2pt}
\setlength{\intextsep}{8pt plus 2pt minus 2pt}

\subsection{Simulation Evaluation}
\label{sec:exp-sim}

We evaluate ABot-N1 in simulation across all five core tasks. To
keep the comparison directly interpretable, every task is reported
under the most widely adopted protocol in its respective literature
(VLN-CE R2R/RxR for \textit{instruction-following}, TrackVLA's EVT-Bench for
\textit{person-following}) or under the closed-loop benchmarks introduced in
Sec.~\ref{sec:benchmark} (ABotN-PointBench, ABotN-POIBench, and an
\textit{object-goal} benchmark derived from OVON, defined below). Unless
specified, ``\textbf{ABot-N1}'' refers to the joint multi-task
checkpoint co-trained on all five tasks---our primary model
demonstrating generalist capability and scaling---and
``ABot-N1$^\dagger$'' to the variant fine-tuned on each
individual task, demonstrating the intrinsic framework advantage.
\begin{table}[t]
\centering
\caption{\textbf{Quantitative Instruction-Following Results on the VLN-CE Benchmarks.} Comparison on the VLN-CE Val-Unseen splits of R2R-CE and
RxR-CE. ``S.RGB''\,=\,single-view RGB; ``Pano.''\,=\,panoramic
RGB; ``Depth''/``Odo.'' indicate depth and odometry availability.
Methods marked with $^{\ast}$ use the waypoint predictor
from~\cite{hong2022bridge}.
ABot-N1 achieves the best NE, SR, and SPL on R2R-CE and the best NE on RxR-CE, using only panoramic RGB without depth or odometry.}
\label{tab:exp-vlnce}
\small
\setlength{\tabcolsep}{3pt}
\begin{tabular}{l cccc cccc ccc}
\toprule
 & \multicolumn{4}{c}{\textbf{Observation}} & \multicolumn{4}{c}{\textbf{R2R-CE Val-Unseen}} & \multicolumn{3}{c}{\textbf{RxR-CE Val-Unseen}} \\
\cmidrule(lr){2-5} \cmidrule(lr){6-9} \cmidrule(lr){10-12}
\textbf{Method} & S.RGB & Pano. & Depth & Odo. & NE$\downarrow$ & OSR$\uparrow$ & SR$\uparrow$ & SPL$\uparrow$ & NE$\downarrow$ & SR$\uparrow$ & SPL$\uparrow$ \\
\midrule
HPN+DN$^{\ast}$~\cite{krantz2021waypoint}          &            & \checkmark & \checkmark & \checkmark & 6.31 & 40.0 & 36.0 & 34.0 & --    & --   & --   \\
CMA$^{\ast}$~\cite{hong2022bridge}                 &            & \checkmark & \checkmark & \checkmark & 6.20 & 52.0 & 41.0 & 36.0 & 8.76  & 26.5 & 22.1 \\
Sim2Sim$^{\ast}$~\cite{krantz2022sim2sim}          &            & \checkmark & \checkmark & \checkmark & 6.07 & 52.0 & 43.0 & 36.0 & 8.76  & 26.5 & 22.1 \\
GridMM$^{\ast}$~\cite{wang2023gridmm}              &            & \checkmark & \checkmark & \checkmark & 5.11 & 61.0 & 49.0 & 41.0 & --    & --   & --   \\
DreamWalker$^{\ast}$~\cite{wang2023dreamwalker}    &            & \checkmark & \checkmark & \checkmark & 5.53 & 59.0 & 49.0 & 44.0 & --    & --   & --   \\
Reborn$^{\ast}$~\cite{an2022rxrhabitat}            &            & \checkmark & \checkmark & \checkmark & 5.40 & 57.0 & 50.0 & 46.0 & 5.98  & 48.6 & 42.0 \\
ETPNav$^{\ast}$~\cite{an2024etpnav}                &            & \checkmark & \checkmark & \checkmark & 4.71 & 65.0 & 57.0 & 49.0 & 5.64  & 54.7 & 44.8 \\
HNR$^{\ast}$~\cite{wang2024lookahead}                    &            & \checkmark & \checkmark & \checkmark & 4.42 & 67.0 & 61.0 & 51.0 & 5.50  & 56.3 & 46.7 \\
\midrule
InstructNav~\cite{long2024instructnav}             &            & \checkmark & \checkmark & \checkmark & 6.89 & --   & 31.0 & 24.0 & --    & --   & --   \\
AO-Planner~\cite{chen2025affordances}                &            & \checkmark & \checkmark &            & 5.55 & 59.0 & 47.0 & 33.0 & --    & --   & --   \\
NaVid~\cite{zhang2024navid}                        & \checkmark &            &            &            & 5.47 & 49.0 & 37.0 & 35.0 & --    & --   & --   \\
Uni-NaVid~\cite{zhang2024uninavid}                 & \checkmark &            &            &            & 5.58 & 53.5 & 47.0 & 42.7 & 6.24  & 48.7 & 40.9 \\
NaVILA~\cite{cheng2025navila}                      & \checkmark &            &            &            & 5.22 & 62.5 & 54.0 & 49.0 & 6.77  & 49.3 & 44.0 \\
StreamVLN~\cite{wei2025streamvln}                  & \checkmark &            &            &            & 4.98 & 64.2 & 56.9 & 51.9 & 6.22  & 52.9 & 46.0 \\
CorrectNav~\cite{yu2026correctnav}                & \checkmark &            &            &            & 4.24 & 67.5 & 65.1 & 62.3 & 4.09  & 69.3 & 63.3 \\
NavFoM~\cite{zhang2025navfom}                      &            & \checkmark &            &            & 4.61 & 72.1 & 61.7 & 55.3 & 4.74  & 64.4 & 56.2 \\
NavForesee~\cite{liu2025navforesee}                &            & \checkmark &            &            & 3.94 & \textbf{78.4} & 66.2 & 59.7 & 4.20  & 66.3 & 53.2 \\

Qwen-VLA~\cite{wang2026qwenvla}                    & \checkmark &            &            &            & 5.10 & 69.0 & 57.3 & 51.2 & 5.80  & 59.6 & 47.8 \\
ABot-N0~\cite{abotn0}                              &            & \checkmark &            &            & 3.80 & 70.8 & 66.4 & 63.9 & 3.83 & 69.3 & 60.0 \\
Qwen-RobotNav-4B \cite{zhang2026qwenrobotnav}                    &            & \checkmark &            &            & 3.80 & 77.2 & 69.5 & 63.6 & --            & \textbf{75.2} & \textbf{65.0} \\
\midrule
ABot-N1$^\dagger$            &            & \checkmark &            &            & 3.91 & 71.7 & 68.3 & 66.6 & 3.43    & 70.9   & 61.4   \\
\textbf{ABot-N1}          &            & \checkmark &            &            & \textbf{3.32} & 75.2 & \textbf{70.9} & \textbf{67.5} & \textbf{3.13}     & 73.9   & 63.9   \\
\bottomrule
\end{tabular}
\end{table}

\FloatBarrier
\subsubsection{Instruction-Following: VLN-CE R2R/RxR}
\label{sec:exp-vlnce}

\paragraph{Protocol.}
We evaluate on the Habitat~\cite{savva2019habitat} VLN-CE
Val-Unseen splits of R2R-CE~\cite{krantz2020vlnce} and
RxR-CE~\cite{ku2020rxr}, using standard metrics: navigation
error (NE), oracle success rate (OSR), success rate
(SR), and success weighted by path length (SPL).
Table~\ref{tab:exp-vlnce} categorizes methods to provide a clear
comparison for ABot-N1:
(i)~sensor-heavy methods (top block) using panoramic RGB, depth,
and odometry-based waypoints;
(ii)~recent foundation models and VLMs (middle block) that shift
towards minimal sensor suites, including both single-view (S.RGB)
and panoramic (Pano.) inputs without relying on specialized mapping;
and (iii)~our proposed ABot-N1 (bottom block). Note that 
ABot-N1 takes only a front-left-right tri-view---three
fixed camera views---rather than a dense $360^\circ$ panoramic sweep,
which we categorize under the panoramic (Pano.) column for brevity.

\paragraph{Analysis.}
ABot-N1 sets a new state of the art on R2R-CE Val-Unseen in NE
(3.32\,m), SR (70.89\%), and SPL
(67.5\%), surpassing all baselines---including
depth-and-odometry methods in the top block---while using only a
front-left-right tri-view of RGB. NavForesee retains the highest OSR (78.4\%); ABot-N1
trades marginal OSR for substantially higher SR/SPL, indicating more
decisive goal-reaching rather than merely passing near the target.
On RxR-CE Val-Unseen, ABot-N1 achieves the best NE (\textbf{3.13\,m})
across all reported methods. SR (73.85\%) and SPL (63.85\%) rank second
only to Qwen-RobotNav-4B (75.2\%/65.0\%), which was trained on
significantly larger in-domain RxR data; ABot-N1 nonetheless
outperforms all other methods including CorrectNav (69.3\%/63.3\%)
and NavFoM (64.4\%/56.2\%).
Comparing ABot-N1 with ABot-N1$^\dagger$: the joint multi-task model
matches or exceeds the single-task specialist on both splits (R2R SR
+2.63, RxR SR +2.95), demonstrating that multi-task co-training
yields positive cross-task transfer rather than catastrophic
interference.
We attribute the gain over prior work to two mechanisms: the
affordance pixel re-grounds the slow system's instruction-progress
reasoning into image-space anchors that the fast controller
tracks, and the explicit
``done\,/\,current'' sub-instruction CoT eliminates the
ambiguous mid-segment behavior that single-system VLN policies often
exhibit on long instructions.

\FloatBarrier
\subsubsection{Object-Goal: Short-Horizon OVON}
\label{sec:exp-objgoal}

\paragraph{Protocol.}
As motivated in Sec.~\ref{sec:bench-pointgoal} and the
Introduction, naive long-horizon \textit{object-goal} evaluation conflates
``find the object'' with ``reach the object'', confounding
recognition errors with execution failures. We therefore evaluate
on a re-curated short-horizon OVON benchmark \cite{yokoyama2024hmovon}: starting from the
original OVON Val-Unseen split, we manually annotate the first
frame in which each target object becomes visible across 36
scenes, and re-anchor the start pose to that frame. The resulting
episodes isolate the recognition-and-approach phase, providing a
clean measurement of out-of-vocabulary generalization and
final-meters arrival. We report SR, SPL, and the average
Distance-to-Goal (DTG) at termination.

\begin{table}[t]
\centering
\caption{\textbf{Quantitative Object-Goal Results on the Short-Horizon OVON Benchmark.} 
We report the SR, SPL, and DTG at episode termination. 
Our proposed \textbf{ABot-N1} establishes new state-of-the-art (SOTA) performance by substantially improving both SR and SPL over existing baselines, while successfully halving the final distance-to-goal (DTG).}

\label{tab:exp-objgoal}
\small
\begin{tabular}{lccc}
\toprule
\textbf{Method} & SR$\uparrow$ & SPL$\uparrow$ & DTG$\downarrow$ \\
\midrule
StreamVLN \cite{wei2025streamvln}              & 39.7 & 15.8 & 2.368 \\
NaVILA    \cite{cheng2025navila}             & 55.4 & 26.1 & 1.811 \\
InternVLA-N1 (S2 only) \cite{wei2025ground} & 58.5 & 30.0 & 1.441 \\
Uni-NaVid \cite{zhang2024uninavid}            & 68.7 & 34.5 & 1.495 \\
ABot-N0 \cite{abotn0}               & 73.2 & 35.4 & 1.442 \\
\midrule
ABot-N1$^\dagger$            & \textbf{85.5} & 50.8 & \textbf{0.592} \\
\textbf{ABot-N1}          & 84.9 & \textbf{51.8} & 0.822 \\
\bottomrule
\end{tabular}
\end{table}

\paragraph{Analysis.}
ABot-N1 raises SR by +11.7 points and SPL by +16.4 points over
ABot-N0, while compressing DTG from 1.44\,m to 0.82\,m---a
1.8$\times$ reduction reflecting improved last-meter precision.
The single-task ABot-N1$^\dagger$ pushes SR slightly higher (85.5)
and DTG lower (0.59\,m), demonstrating the framework's raw per-task
capability. ABot-N1 only marginally trails ABot-N1$^\dagger$ in SR
(by 0.6 points) while in fact \emph{exceeding} the specialist on
SPL (51.8 vs.\ 50.8), evidencing strong cross-task transfer rather
than catastrophic interference.
The gap to single-system baselines (Uni-NaVid, NaVILA) widens
most on DTG and SPL, consistent with our hypothesis that
decoupling recognition (object pixel) from approach (affordance
pixel) removes the exploration--arrival entanglement that
bottlenecks monolithic policies on out-of-vocabulary targets.

\FloatBarrier
\subsubsection{Point-Goal: ABotN-PointBench}
\label{sec:exp-pointgoal}

\paragraph{Protocol.}
We evaluate on ABotN-PointBench (Sec.~\ref{sec:bench-pointgoal}) in closed-loop rollouts within our in-house 3DGS evaluator, separating outdoor and indoor splits and further stratifying each into difficulty tiers (outdoor: Low / Medium / High; indoor: Low / High) based on path length and environmental complexity. Outdoor metrics are SR and SPL. The outdoor split adopts a relaxed collision criterion suited to complex open environments: an episode is deemed successful only if the agent arrives with fewer than three collision events (SR$_{<3\text{col}}$). This redundant collision budget accounts for the slight annotation deviations inherent in the large-scale outdoor occupancy maps, preventing minor labeling inaccuracies near obstacle boundaries from being mistaken for genuine navigation failures. The indoor split adopts a strict collision criterion: an episode is deemed successful only if the agent arrives with no collision event.

Importantly, the success rate (SR) serves as a holistic indicator that naturally reflects the agent's social compliance over both distance and time. Since a socially non-compliant policy (e.g., one that frequently cuts across vehicle lanes or enters restricted pedestrian zones) inevitably fails to navigate safely, achieving a successful, collision-free arrival inherently ensures that the agent has spent the vast majority of its travel distance and execution time within socially compliant pathways.

\begin{table}[t]
\centering
\caption{\textbf{Quantitative Point-Goal Results on the Outdoor Split of ABotN-PointBench.} 
We report the success rate under a redundant collision budget ($\text{SR}_{<3\text{col}}$) and SPL, stratified by difficulty tiers: Low (L), Medium (M), and High (H). 
Our proposed \textbf{ABot-N1} establishes new SOTA performance by consistently securing the highest SR and SPL across all difficulty levels, demonstrating its superior capacity for safe and socially compliant navigation in complex open environments.}

\label{tab:exp-pointgoal-out}
\small
\setlength{\tabcolsep}{6.0pt}
\begin{tabular}{l cccc cccc}
\toprule
 & \multicolumn{4}{c}{\textbf{SR$_{<3\text{col}}\uparrow$}} & \multicolumn{4}{c}{\textbf{SPL$\uparrow$}} \\
\cmidrule(lr){2-5} \cmidrule(lr){6-9}
\textbf{Method} & All & L & M & H & All & L & M & H \\
\midrule
GNM~\cite{shah2023gnm}                & 39.1 & 66.7 & 32.0 & 18.7 & 36.7 & 65.3 & 28.4 & 16.5 \\
ViNT~\cite{shah2023vint}              & 62.2 & 92.0 & 50.7 & 44.0 & 62.2 & 92.0 & 50.6 & 44.0 \\
NoMaD~\cite{sridhar2024nomad}         & 56.0 & 90.7 & 45.3 & 32.0 & 55.7 & 90.0 & 45.3 & 31.6 \\
CityWalker~\cite{liu2025citywalker}   & 48.9 & 73.3 & 40.0 & 33.3 & 48.3 & 72.2 & 39.7 & 32.8 \\
SocialNav~\cite{chen2025socialnav}    & 72.0 & 93.3 & 64.0 & 58.7 & 71.9 & 93.2 & 63.9 & 58.7 \\
ABot-N0~\cite{abotn0}                 & 76.9 & 93.3 & 72.0 & 65.3 & 76.9 & 93.3 & 72.0 & 65.3 \\
\midrule
ABot-N1$^\dagger$                     & 92.0 & \textbf{98.7} & 90.7 & 86.7 & 90.7 & \textbf{96.7} & 89.4 & 85.8 \\
\textbf{ABot-N1}                      & \textbf{92.9} & 96.0 & \textbf{94.7} & \textbf{88.0} & \textbf{91.4} & 95.6 & \textbf{92.3} & \textbf{86.4} \\
\bottomrule
\end{tabular}
\end{table}

\begin{table}[t]
\centering
\caption{\textbf{Quantitative Point-Goal Results on the Indoor Split of ABotN-PointBench.} 
We report the success rate under a strict zero-collision criterion ($\text{SR}_{<1\text{col}}$) and SPL, stratified by difficulty tiers: Low (L) and High (H). 
Our proposed \textbf{ABot-N1} establishes new SOTA performance by achieving the highest SR ($95.4\%$) and SPL across all difficulty levels, demonstrating its exceptional safety and precision in constrained indoor environments.}

\label{tab:exp-pointgoal-in}
\small
\setlength{\tabcolsep}{6.0pt}
\begin{tabular}{l ccc ccc}
\toprule
 & \multicolumn{3}{c}{\textbf{SR$_{<1\text{col}}\uparrow$}} & \multicolumn{3}{c}{\textbf{SPL$\uparrow$}} \\
\cmidrule(lr){2-4} \cmidrule(lr){5-7}
\textbf{Method} & All & L & H & All & L & H \\
\midrule
GNM~\cite{shah2023gnm}                & 26.7 & 30.8 & 22.5 & 26.6 & 30.6 & 22.5 \\
ViNT~\cite{shah2023vint}              & 27.9 & 32.5 & 23.3 & 27.9 & 32.4 & 23.3 \\
NoMaD~\cite{sridhar2024nomad}         & 20.0 & 20.8 & 19.2 & 19.6 & 20.5 & 18.6 \\
CityWalker~\cite{liu2025citywalker}   & 21.7 & 23.3 & 20.0 & 21.6 & 23.3 & 19.9 \\
SocialNav~\cite{chen2025socialnav}    & 42.5 & 46.7 & 38.3 & 42.5 & 46.7 & 38.3 \\
ABot-N0~\cite{abotn0}                 & 89.6 & 91.7 & 87.5 & 85.6 & 87.5 & 83.7 \\
\midrule
ABot-N1$^\dagger$                     & 90.4 & 92.5 & 88.3 & 89.4 & 91.5 & 87.4 \\
\textbf{ABot-N1}                      & \textbf{95.4} & \textbf{95.0} & \textbf{95.8} & \textbf{93.7} & \textbf{93.3} & \textbf{94.2} \\
\bottomrule
\end{tabular}
\end{table}

\paragraph{Analysis.}
Outdoor results highlight the structural advantage of the
slow-fast factorization: ABot-N1 reaches 92.9\% overall SR with
strong social compliance, improving over ABot-N0
by +16.0 SR while simultaneously raising compliance performance.
The difficulty-stratified view reveals that the gain is particularly
pronounced on the hardest tier (SR: 88.0 vs.\ 65.3, a +22.7-point
jump), where prior policies---lacking any notion of traversable
regions---simply follow the geometrically shortest path to the goal
coordinate, cutting across vehicle lanes and restricted regions and
thus failing to maintain safety and compliance.
By generating a CoT-guided affordance pixel that re-grounds
navigation onto the nearest traversable pathway at every step,
ABot-N1 recovers safe, compliant trajectories that a metric-only
policy cannot maintain.
ABot-N1$^\dagger$ (92.0\% SR / 90.7\% SPL) already outperforms
all baselines, confirming the framework advantage independent of
multi-task data; the full ABot-N1 further improves all metrics,
especially on Medium and High tiers, suggesting that social-rule
semantics learned via \textit{object-goal} and \textit{POI-goal} supervision transfer
back into \textit{point-goal} compliant behavior.
Indoor performance under the strict no-collision criterion
shows ABot-N1 at 95.4\% SR---a +5.8-point gain over ABot-N0
(89.6\%)---with the improvement concentrated on the High-difficulty
set (95.8 vs.\ 87.5), indicating robust generalization to complex
interior layouts.

\FloatBarrier
\subsubsection{POI-Goal: ABotN-POIBench}
\label{sec:exp-poigoal}

\paragraph{Protocol.}
We evaluate on ABotN-POIBench (Sec.~\ref{sec:bench-poigoal}). The
success criterion is final-meters arrival to the annotated
entrance frame: an episode is successful if the agent terminates
within a 2\,m radius of the entrance (SR$_{<2\mathrm{m}}$). We
additionally report SPL against the A$^{\!*}$ reference
trajectory.

\begin{table}[t]
\centering
\caption{\textbf{Quantitative POI-Goal Results on the Urban-Scale ABotN-POIBench.} 
We report the entrance arrival success rate ($\text{SR}_{<2\mathrm{m}}$) and SPL compared against representative baselines. 
Our proposed \textbf{ABot-N1} establishes a new SOTA, boosting the success rate to \textbf{77.3\%} and achieving \textbf{72.6\%} SPL.}


\label{tab:exp-poigoal}
\small
\begin{tabular}{lcc}
\toprule
\textbf{Method} & SR$_{<2\mathrm{m}}\uparrow$ & SPL$\uparrow$ \\
\midrule
ViNT \cite{shah2023vint}                        & 19.0 & 18.2 \\
OmniNav (vanilla) \cite{xue2025omninav}           & 23.9 & 22.4 \\
OmniNav (BridgeNav training)\cite{xue2025omninav} & 34.4 & 31.5 \\
ABot-N0   \cite{abotn0}                   & 20.9 & 17.7 \\
POINav~\cite{gong2026poinav} & 42.3 & 40.3 \\
\midrule
ABot-N1$^\dagger$             & 69.9 & 59.8 \\
\textbf{ABot-N1}           & \textbf{77.3} & \textbf{72.6} \\
\bottomrule
\end{tabular}
\end{table}

\paragraph{Analysis.}
ABot-N1 sets a new state of the art on ABotN-POIBench, raising
SR$_{<2\mathrm{m}}$ from 42.3\% (POINav) to 77.3\% (+35.0 points)
and SPL from 40.3\% to 72.6\% (+32.3 points). Even ABot-N1$^\dagger$
variant already achieves 69.9\%\,/\,59.8\%, confirming the framework
advantage alone; the full ABot-N1 pushes further, benefiting from
the largest cross-task synergy we observe: \textit{POI-goal} supervises
only the slow system, so its fast controller benefits from
\textit{point-goal} data, while its slow system benefits from
\textit{object-goal}-style visibility reasoning. The contrast with
OmniNav variants is instructive---OmniNav (BridgeNav training)
improves over the vanilla version by 10 points but plateaus,
consistent with the documented BridgeNav rollout-bias on
large-heading-offset starts; ABot-N1's slow system, trained on
$\sim$3.0\,M tri-view annotations including 0.5\,M hard negatives,
avoids this trajectory bias by construction.

\FloatBarrier
\subsubsection{Person-Following: EVT-Bench}
\label{sec:exp-pf}

\paragraph{Protocol.}
We evaluate \textit{person-following} on the closed-loop
EVT-Bench~\cite{wang2025trackvla} in Habitat, using the
identical protocol as ABot-N0~\cite{abotn0} and
TrackVLA~\cite{wang2025trackvla}: each episode requires
the agent to follow a target person under three difficulty
levels -- Single-Target Tracking (STT, nominal following),
Distracted Tracking (DT, similar-looking pedestrians causing
identity confusion), and Ambiguity Tracking (AT, frequent
severe occlusions). For each scenario we report Success
Rate (SR), Tracking Rate (TR, formerly ``following rate''),
and Collision Rate (CR), all in percent.

\begin{table}[t]
\centering
\caption{\textbf{Quantitative Person-Following Results on EVT-Bench~\cite{wang2025trackvla}.}
SR/TR in \%, higher is better; CR in \%, lower is better. The
symbols $\dagger$ and $\ddagger$ indicate the adoption of
GroundingDINO and the SoM+GPT-4o pipeline, respectively, as in
ABot-N0~\cite{abotn0}.
ABot-N1 achieves the best SR and TR across the three splits with the sole exception of STT TR, surpassing ABot-N0 and all baselines.}
\label{tab:exp-pf}
\setlength{\tabcolsep}{3pt}
\small
\begin{tabular}{l ccc ccc ccc}
\toprule
 & \multicolumn{3}{c}{\textbf{Single-Target (STT)}} & \multicolumn{3}{c}{\textbf{Distracted (DT)}} & \multicolumn{3}{c}{\textbf{Ambiguity (AT)}} \\
\cmidrule(lr){2-4} \cmidrule(lr){5-7} \cmidrule(lr){8-10}
\textbf{Method} & SR$\uparrow$ & TR$\uparrow$ & CR$\downarrow$ & SR$\uparrow$ & TR$\uparrow$ & CR$\downarrow$ & SR$\uparrow$ & TR$\uparrow$ & CR$\downarrow$ \\
\midrule
IBVS$^{\ddagger}$~\cite{gupta2016novel}    & 42.9 & 56.2 & 3.75 & 10.6 & 28.4 & 6.14 & 15.2 & 39.5 & \textbf{4.90} \\
PoliFormer$^{\dagger}$~\cite{zeng2024poliformer} & 4.67 & 15.5 & 40.1 & 2.62 & 13.2 & 44.5 & 3.04 & 15.4 & 41.5 \\
EVT$^{\ddagger}$~\cite{zhong2024empowering}      & 32.5 & 49.9 & 40.5 & 15.7 & 35.7 & 53.3 & 18.3 & 21.0 & 44.9 \\
Uni-NaVid~\cite{zhang2024uninavid}            & 25.7 & 39.5 & 41.9 & 11.3 & 27.4 & 43.5 & 8.26 & 28.6 & 43.7 \\
NavFoM~\cite{zhang2025navfom}                 & 85.0 & 80.5 & --   & 61.4 & 68.2 & --   & --   & --   & --   \\
TrackVLA++~\cite{wang2025trackvlapp}          & 86.0 & 81.0 & \textbf{2.10} & 66.5 & 68.8 & \textbf{4.71} & 51.2 & 63.4 & 15.9 \\
Qwen-RobotNav-4B~\cite{zhang2026qwenrobotnav}                              & 77.4 & \textbf{90.0} & 6.40 & --   & --   & --   & --   & --   & --   \\
Qwen-RobotNav-8B~\cite{zhang2026qwenrobotnav}                                & 78.6 & 89.7 & 5.70 & --   & --   & --   & --   & --   & --   \\
ABot-N0~\cite{abotn0}                         & 86.9 & 87.6 & 8.54 & 66.7 & 75.4 & 11.6 & 67.3 & 79.5 & 7.05 \\
\midrule
ABot-N1$^\dagger$                      & 87.0 & 86.9 & 6.83 & 65.2 & 72.7 & 14.7 & \textbf{70.3} & 81.1 & 9.47 \\
\textbf{ABot-N1}                       & \textbf{90.1} & 89.8 & 4.27 & \textbf{67.4} & \textbf{84.4} & 17.9 & 70.0 & \textbf{87.8} & 17.9 \\
\bottomrule
\end{tabular}
\end{table}

\paragraph{Analysis.}
ABot-N1 leads on all SR metrics and on TR for the DT and AT
splits, while matching the top STT TR within 0.2 points. On the easy STT setting, it reaches 90.1 SR /
89.8 TR---surpassing ABot-N0 by +3.2 SR and +2.2 TR while
maintaining comparable CR (4.27 vs.\ 8.54). On the hardest AT split,
ABot-N1 achieves 70.0 SR / 87.8 TR,
improving over ABot-N0 by +2.7 SR and +8.3 TR, confirming that the
asymmetric pixel protocol (target-pixel set to null while the
affordance pixel falls back to the first projectable future waypoint,
cf.\ Sec.~\ref{sec:data-tasks}) effectively hardens re-identification
across severe occlusions. The most striking gain appears on DT, where
ABot-N1 raises TR from 75.4 to 84.4 (+9.0 points) and
SR from 66.7 to 67.4, indicating that joint multi-task
training strengthens the slow system's identity reasoning even
under distractor interference.
The trade-off is a slightly higher CR on DT/AT (17.9\% vs.\ 11.6\%
/ 7.05\%), reflecting a more aggressive approach policy learned from
the combined training set; future work may mitigate this via an explicit collision-avoidance reward during post-training.

\FloatBarrier
\subsubsection{Cross-Task Summary}
\label{sec:exp-summary}

Across the five tasks evaluated in this report,
ABot-N1 improves over ABot-N0 on the primary metric of
every task: +3.9 SR on R2R-CE,
+16.0 SR on ABotN-PointBench (outdoor split) with simultaneously improved
compliance, +0.558 absolute jump in SR$_{<2\mathrm{m}}$ on
ABotN-POIBench, and +3.2 STT-SR / +2.7 AT-SR on EVT-Bench
Person-Following. On short-horizon OVON, ABot-N1$^\dagger$
variant lifts SR by +12.3, confirming the framework advantage
even without multi-task data.
ABot-N1 matches or exceeds ABot-N1$^\dagger$
on \textit{point-goal}, \textit{instruction-following}, and \textit{POI-goal} and only
marginally trails on \textit{object-goal}, indicating that the
unified pixel-goal interface supports positive cross-task
transfer and effective scaling rather than the destructive
interference often reported when training generalist navigators
on heterogeneous data.

\subsection{Real-World Deployment}
\label{sec:exp-real}

To validate the practical applicability of the proposed system, we deploy
the complete ABot-N1 on our proprietary AMap TuTu
quadrupedal robot platform and conduct extensive real-world experiments
in diverse indoor and outdoor environments.

\subsubsection{Hardware Setting}
\label{sec:exp-hw}

The experimental platform is \textbf{AMap TuTu}, a 12-DOF quadrupedal robot equipped with omnidirectional LiDAR, RTK-GNSS, and a Tri-Camera array spanning $270^{\circ}$ horizontal FOV for egocentric perception. 
For ABot-N1, we introduce two key upgrades.
First, we deeply integrate the previously modular perception stack (BEV occupancy mapping, legged odometry, and obstacle filtering) with the VLN inference pipeline into a single unified onboard
framework, eliminating inter-process communication overhead and
reducing end-to-end latency.
Second, we upgrade the compute module from Jetson Orin NX to
Jetson AGX Orin 64\,GB. The $4\times$ memory
expansion and increased GPU throughput are critical for hosting the
dual slow-fast VLN inference concurrently with real-time BEV
perception and locomotion control, enabling the full
navigation architecture to run entirely on-robot without offloading.

\subsubsection{Edge Deployment Scheme}
\label{sec:exp-deploy}

For real-time execution on the edge device, we deploy the full
slow-fast pipeline at an overall control frequency of 10\,Hz.
To ensure robust performance under limited computational resources,
a lightweight model structure together with an asynchronized
sequential execution scheme are deliberately designed as follows.

To reduce parameter size, the slow system is built upon
Qwen-3.5-2B, while the fast system is implemented by a DiT with
306\,M parameters whose visual observations are encoded by a
DINOv2-Base encoder. The two systems operate asynchronously: the
slow system runs at a lower frequency than the fast system, whereas
the fast system runs continuously with cached information from the
slow system and current observations over multiple control cycles.
Once the slow system produces an updated plan, the fast system
generates waypoints conditioned on the latest reasoning result.
This decoupled scheduling maximizes GPU utilization on the Jetson AGX Orin
while maintaining consistent 10\,Hz closed-loop responsiveness.


\subsubsection{Deployment Results}
\label{sec:exp-results}

We present qualitative deployment results across the five core tasks,
demonstrating ABot-N1's real-world reasoning and control capabilities
through representative episodes with slow-system CoT and pixel-level
prediction visualizations.
\begin{figure}[!t]
\centering
\includegraphics[width=\linewidth]{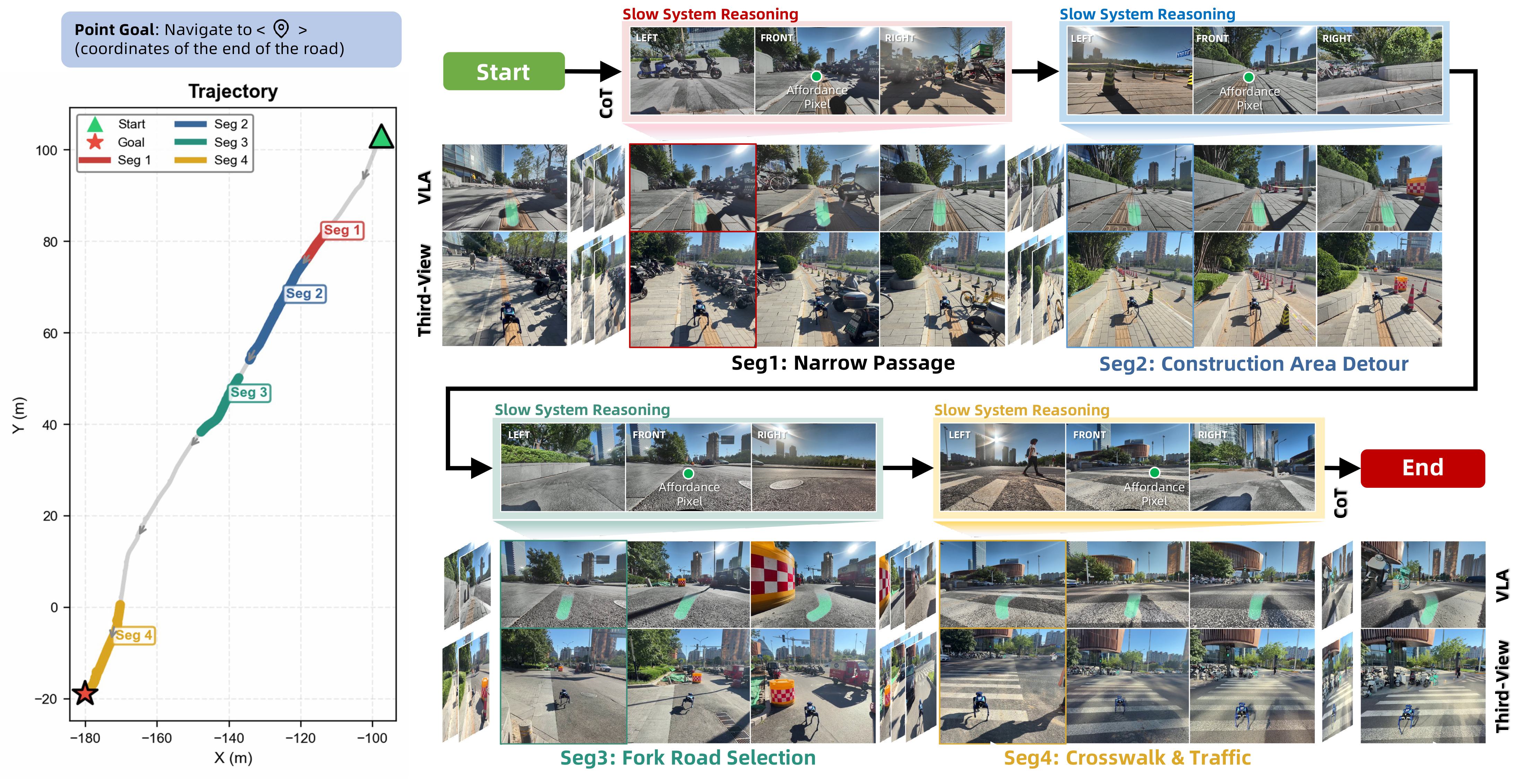}
\caption{\textbf{Point-Goal Deployment.} Four segments of a long-range outdoor
episode showcasing obstacle avoidance on narrow roads, construction area detour, correct fork selection, and traffic-light-compliant crosswalk
traversal.}
\label{fig:exp-point}
\end{figure}

\paragraph{Point-Goal Navigation.}
Figure~\ref{fig:exp-point} illustrates a long-distance outdoor
navigation episode decomposed into four representative segments.
(1)~\emph{Narrow-road obstacle avoidance}: the agent successfully
navigates around parked vehicles on a constricted roadway, with the
affordance pixel consistently anchored to the traversable lane.
(2)~\emph{Construction-zone detour}: upon detecting a
construction area, the slow system places the affordance pixel on a
detour around the restricted region, steering the fast controller
clear of the illegal construction zone rather than entering it.
(3)~\emph{Fork selection}: at a road bifurcation, the affordance pixel
is anchored to the target-aligned branch, steering
the agent onto the correct path and avoiding wrong-turn failures.
(4)~\emph{Traffic-light compliance}: approaching a signalized
crosswalk, the agent waits for the green signal before crossing,
demonstrating awareness of traffic rules and the ability to
temporally gate its own actions---a behavior absent from
metric-only \textit{point-goal} policies.
These results confirm that the slow system's traversability-aware
pixel guidance translates into safe, rule-compliant outdoor navigation
beyond simple goal-reaching.

\begin{figure}[!t]
\centering
\includegraphics[width=\linewidth]{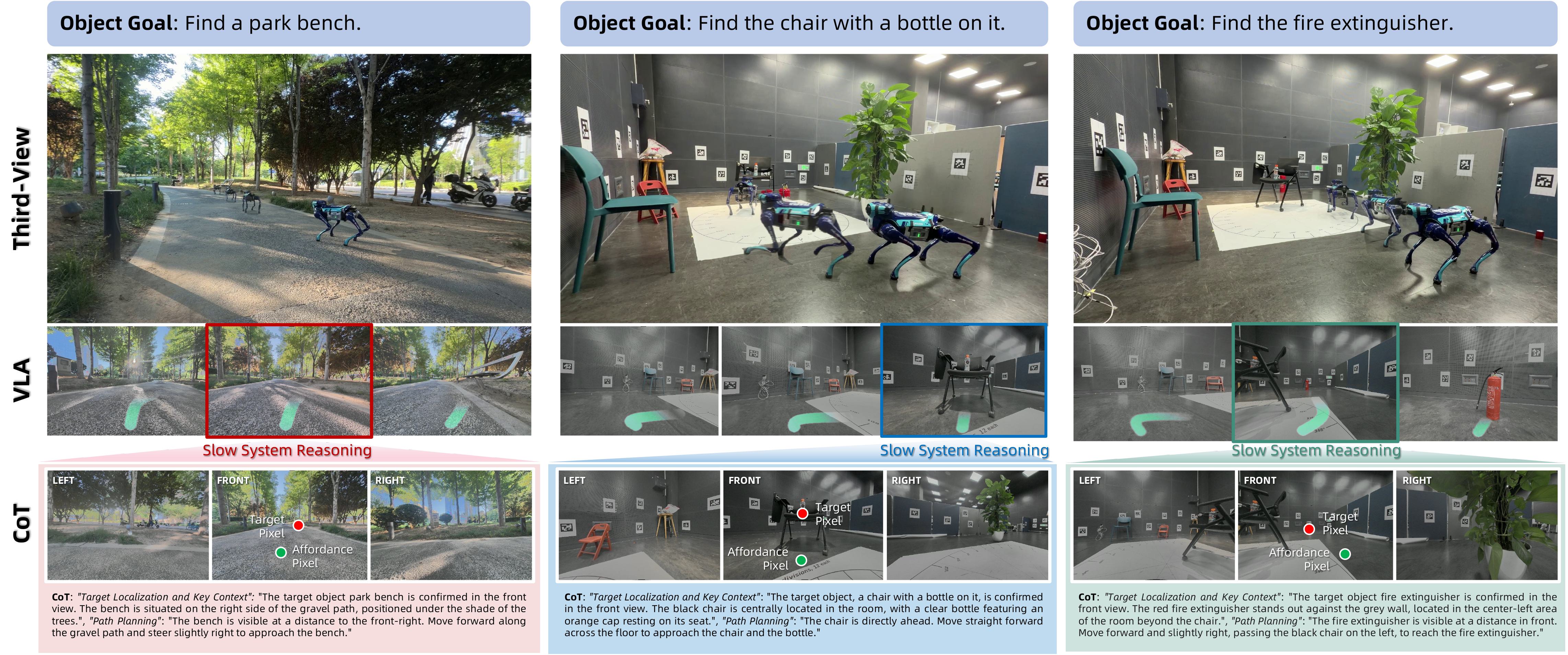}

\caption{\textbf{Object-Goal Deployment.} Three cases---outdoor bench under
dappled tree shade at long range, indoor chair with a water bottle
(spatial reasoning), and partially occluded fire extinguisher---with
CoT, affordance, and target pixel overlays.}
\label{fig:exp-obj}
\end{figure}

\paragraph{Object-Goal Navigation.}
Figure~\ref{fig:exp-obj} presents three representative \textit{object-goal}
episodes with CoT, affordance pixel, and target pixel outputs.
(1)~\emph{Outdoor bench finding}: the agent locates a park bench
under dappled tree shade at a considerable distance, demonstrating
robust long-range recognition under complex natural lighting.
(2)~\emph{Indoor chair with water bottle}: the slow system reasons
about spatial relations between objects---distinguishing the target
chair \emph{with a water bottle} from other chairs---evidencing
spatial reasoning and attribute-aware recognition.
(3)~\emph{Indoor fire extinguisher}: the agent navigates around
indoor obstacles and identifies a partially occluded fire
extinguisher, showcasing perception under occlusion.
Across all cases, the affordance pixel consistently falls on safe
traversable surfaces while the target pixel locks onto the
goal object once visible, validating the decoupled
recognition-and-approach protocol.

\FloatBarrier

\begin{figure}[!t]
\centering
\includegraphics[width=\linewidth]{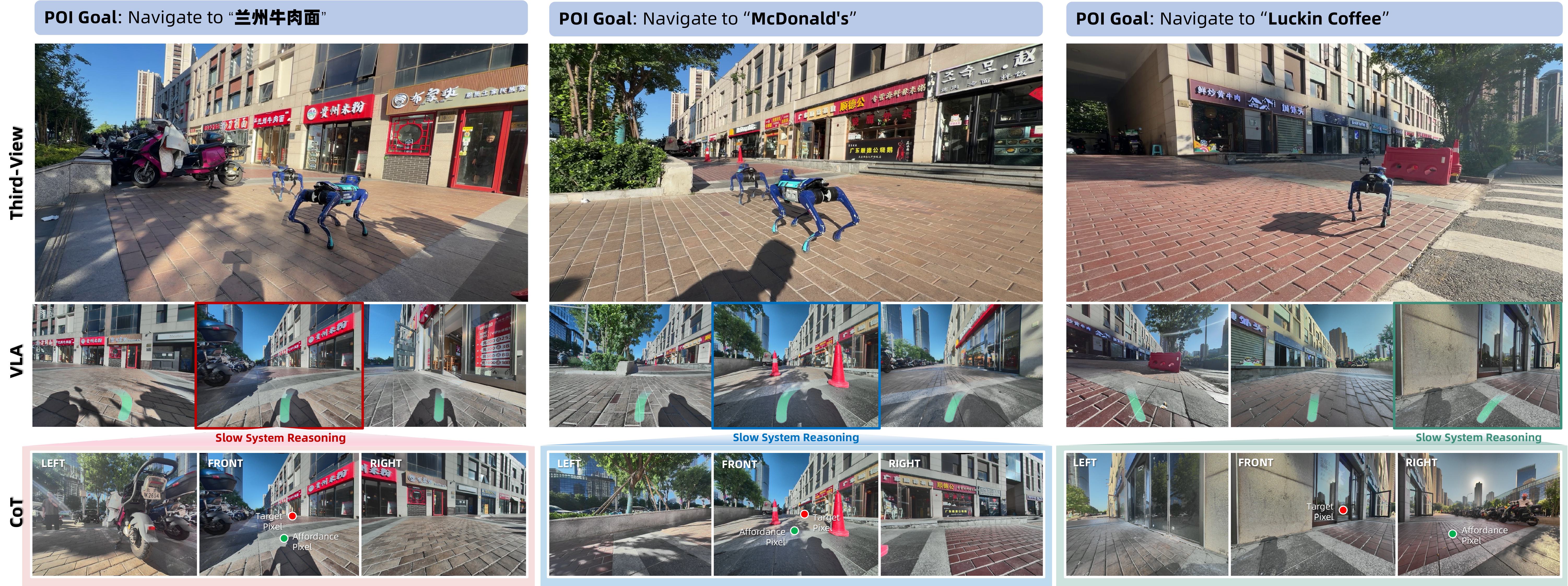}

\caption{\textbf{POI-Goal Deployment.} Locating a Lanzhou noodle restaurant
(large viewing angle), McDonald's (obstacle avoidance en route),
and Luckin Coffee (slope navigation and staircase avoidance).}
\label{fig:exp-poi}
\end{figure}

\FloatBarrier
\paragraph{POI-Goal Navigation.}
Figure~\ref{fig:exp-poi} shows three \textit{POI-goal} episodes in which
the agent locates commercial storefronts by recognizing signage.
(1)~\emph{Lanzhou beef noodle restaurant}: the agent identifies
the shop sign at a large oblique viewing angle, demonstrating
robust text and logo recognition under perspective distortion.
(2)~\emph{McDonald's}: building upon recognition, the agent
performs obstacle avoidance during approach, coordinating the
affordance pixel around street-level obstructions.
(3)~\emph{Luckin Coffee}: the agent navigates a gentle slope,
identifies a passable entrance, and avoids an impassable staircase
with high steps, demonstrating the slow system's joint reasoning
over accessibility and goal identity.
These results highlight that \textit{POI-goal} requires not only
fine-grained visual recognition but also path-feasibility reasoning,
both of which are naturally captured by the target pixel + affordance pixel
protocol.

\begin{figure}[!t]
\centering
\includegraphics[width=\linewidth]{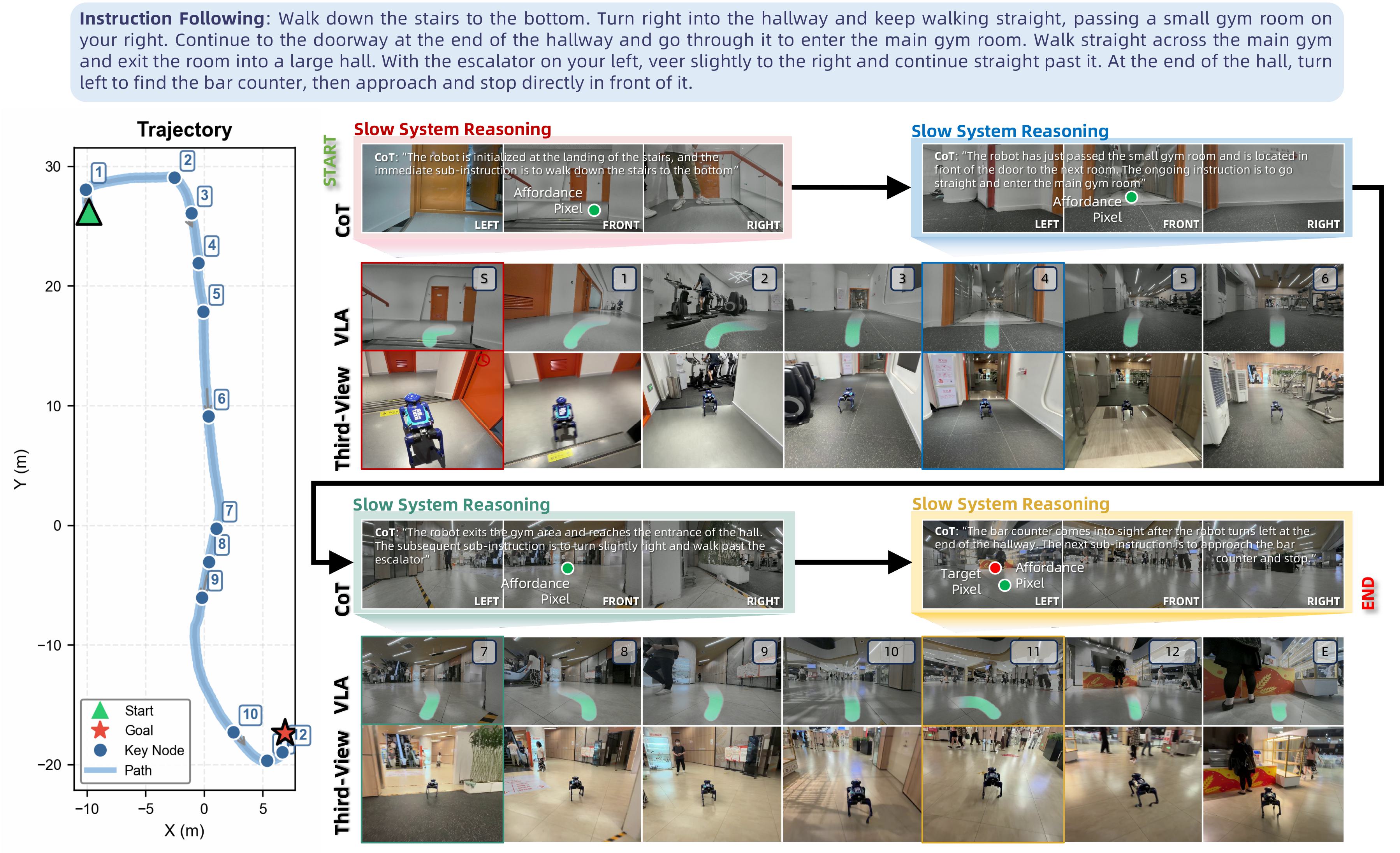}

\caption{ \textbf{Instruction-Following Deployment.} Slow-system reasoning at
four critical moments---stair descent, gym entry, gym exit, and bar
approach---with CoT, affordance pixel, and target pixel
visualizations.}
\label{fig:exp-ins}
\end{figure}

\FloatBarrier
\paragraph{Instruction-Following.}
Figure~\ref{fig:exp-ins} visualizes four key decision moments
during an indoor \textit{instruction-following} episode with the command:
\emph{``Walk down the stairs \ldots\ approach and stop directly in
front of the bar counter.''}  At each moment the slow system
produces a CoT trace and an affordance pixel (and, near the
destination, a target pixel):
(1)~\emph{Descending stairs}: the CoT identifies the current
sub-instruction and the affordance pixel marks the stairwell path.
(2)~\emph{Entering the gym room}: the reasoning correctly advances
the instruction pointer and anchors the pixel to the gym doorway.
(3)~\emph{Exiting the gym}: the CoT tracks cumulative progress and
shifts the pixel toward the hall exit.
(4)~\emph{Approaching the bar counter}: having recognized the
final landmark, the system emits a target pixel directly on the bar
and prepares to terminate.
The explicit CoT decomposes the long instruction into manageable
sub-goals at each step, eliminating the ambiguous mid-segment
behavior typical of single-system VLN policies.

\begin{figure}[!t]
\centering
\includegraphics[width=\linewidth]{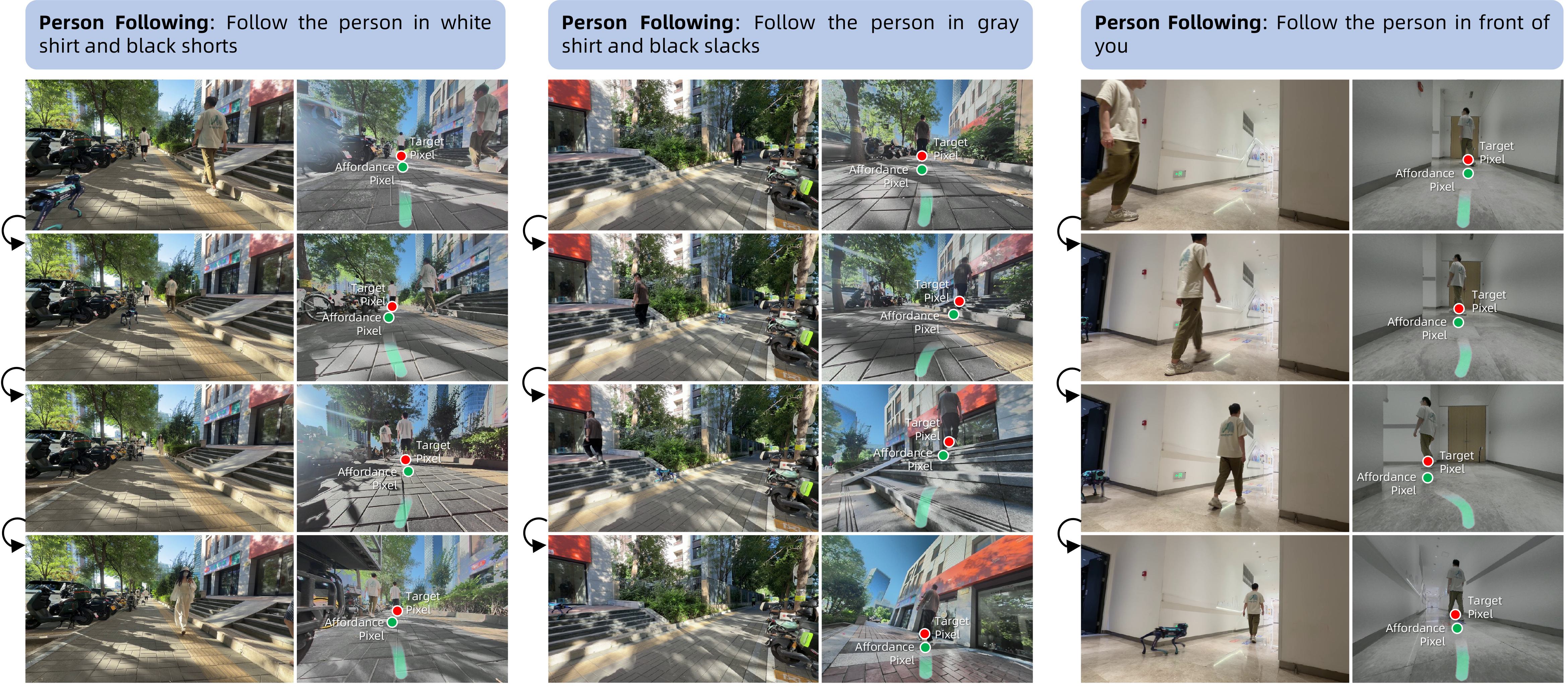}

\caption{\textbf{Person-Following Deployment.} Outdoor tracking under
pedestrian distraction, stair-climbing following, and indoor
corner-rounding with temporary occlusion.}
\label{fig:exp-person}
\end{figure}

\FloatBarrier
\paragraph{Person-Following.}
Figure~\ref{fig:exp-person} illustrates three Person-Following cases.
(1)~\emph{Outdoor following with distractors}: the agent maintains
stable tracking of the designated target despite the presence of
other pedestrians, demonstrating the slow system's robust identity
reasoning and re-identification capability.
(2)~\emph{Outdoor stair-climbing}: the target ascends a staircase
and the agent follows seamlessly, confirming the fast system's
stable low-level control under elevation changes.
(3)~\emph{Indoor corner tracking}: as the target rounds a sharp
indoor corner, the agent adjusts its heading in time and follows
smoothly while keeping the target continuously in view.
The combination of slow-system identity persistence and
fast-system reactive control ensures continuous, collision-aware
following across challenging real-world conditions.

\paragraph{Summary.}
The five deployment tasks collectively validate the slow-fast
architecture: the slow system provides high-level semantic reasoning---instruction
decomposition, open-vocabulary recognition, spatial relation inference,
rule-aware decision-making, and identity persistence---externalized as
CoT traces and pixel-level grounding; the fast system translates these
pixel anchors into smooth  control with precise last-meter arrival,
dynamic obstacle avoidance, and stable locomotion under terrain variation.
Their asynchronous coupling eliminates odometric drift through
continuous visual re-grounding while the fast system's reactivity
compensates for the slow system's latency, yielding robust,
compliant navigation across all five real-world scenarios.

\section{Conclusion}
\label{sec:conclusion}

We have presented ABot-N1, a general visual-language navigation foundation model whose fast--slow architecture factorizes high-level cognition from low-level control: a slow vision-language reasoner produces Chain-of-Thought traces and pixel-level grounding, while a fast action expert converts these into reactive  waypoints. The dual visual-language guidance through affordance and target pixels serves as a universal action interface that accommodates five navigation tasks---\textit{point-goal}, \textit{instruction-following}, \textit{object-goal}, \textit{POI-goal}, and \textit{person-following}---from a single multi-task checkpoint, enabling strong cross-task transfer without task-specific modifications. ABot-N1 achieves state-of-the-art results on all five benchmarks (70.89\% SR on R2R-CE, 92.9\% SR on outdoor \textit{point-goal}, 77.3\% on \textit{POI-goal}, 84.9\% on \textit{object-goal}, and 90.1\% on \textit{person-following}), with the multi-task model consistently matching or exceeding single-task specialists. Real-world deployment on a quadrupedal robot further confirms generalization from simulation to diverse physical environments, demonstrating a viable path toward embodied navigation agents that are simultaneously robust, versatile, and interpretable.

\clearpage

\section{Contributions}
\label{sec:contributions}

\setlength{\parskip}{0pt} 
\setlength{\itemsep}{0pt} 
\setlength{\parsep}{0pt}  
\begin{multicols}{2}
\raggedcolumns

\subsubsection*{Research}
\begin{itemize}
\item Ruiyan Gong
\item Yingnan Guo
\item Junjun Hu
\item Jintao Kong
\item Xiaoxu Leng
\item Tianlun Li
\item Weize Li
\item Fei Liu
\item Zhicheng Liu
\item Jia Lu
\item Minghua Luo
\item Chenlin Ming
\item Yanfen Shen
\item Jiyue Tao
\item Zhengbo Wang
\item Mingyang Yin
\end{itemize}

\subsubsection*{Engineering}
\begin{itemize}
\item Minqi Gu
\item Zihao Guan
\item Wei Guo
\item Guoqing Liu
\item Huachong Pang
\item Menglin Yang
\item Zeqian Ye

\end{itemize}

\columnbreak

\subsubsection*{Data}
\begin{itemize}
\item Xiaoxiao Geng
\item Zhining Gu
\item Honglin Han
\item Di Jing
\item Hongyu Pan
\item Mingchao Sun
\item Kuan Yang
\item Jianfang Zhang

\end{itemize}

\subsubsection*{Infra}
\begin{itemize}
\item Yanghong Chen
\item Ye He
\item Wei Mei
\item Jiahao Shi
\item Xiangpo Yang
\item Yanqing Zhu

\end{itemize}

\subsubsection*{Project Lead}
\begin{itemize}
\item Zedong Chu$^{\dagger}$
\item Xiaolong Wu$^{\dagger}$
\end{itemize}

\subsubsection*{Advisor}
\begin{itemize}
\item Mu Xu
\item Wenbin Tang
\end{itemize}

\end{multicols}

\subsubsection*{Acknowledgments}
We extend our sincere gratitude to the broader team for their support, particularly Wenbin Tang, Zhining Gu, Shihui Su, Zixiao Tang, Yang Cai, Linbo Zheng, and Jingjing Ma.

\vfill
\noindent Authors are listed in alphabetical order.

\noindent $^{\dagger}$ Corresponding authors: chuzedong.czd@alibaba-inc.com, huanlu.wxl@alibaba-inc.com

\clearpage

\clearpage

\bibliographystyle{plainnat}
\bibliography{main}

\end{document}